\documentclass[journal,twoside]{IEEEtran}
\usepackage[utf8]{inputenc}
\usepackage{amsmath,amsfonts}
\usepackage{algorithmic}
\usepackage{array}
\usepackage[export]{adjustbox}
\usepackage{textcomp}
\usepackage{stfloats}
\usepackage{url}
\usepackage{verbatim}
\usepackage{graphicx}
\usepackage{balance}

\usepackage[linesnumbered,ruled]{algorithm2e}
\usepackage{libertine}
\usepackage{hyperref}
\usepackage{cleveref}
\usepackage{todonotes}
\usepackage{multirow}
\usepackage{graphicx}
\usepackage{caption}
\usepackage{csquotes}
\usepackage{bbm}
\usepackage{orcidlink}
\usepackage{pifont}
\usepackage{mathdots}
\usepackage{bm}

\usepackage{booktabs}

\newcommand{\argmax}{\mathop{\mathrm{argmax}}\nolimits} 

\SetKwComment{Comment}{ /*}{ */}

\title{Preference-Learning Emitters for Mixed-Initiative Quality-Diversity Algorithms}

\author{Roberto Gallotta\orcidlink{0000-0001-7578-6173}\thanks{R. Gallotta is a Researcher at Araya Inc., Tokyo, Japan}, Kai Arulkumaran\orcidlink{0000-0003-0459-892X}\thanks{K. Arulkumaran is a Research Team Lead at Araya Inc., Tokyo, Japan}, and L. B. Soros\orcidlink{0000-0002-3259-3205}\thanks{L. B. Soros is a Roman Family Teaching and Research Fellow at Barnard College, New York City, USA and a Researcher at Cross Labs, Cross Compass, Ltd., Kyoto, Japan}
\thanks{\\\copyright 2023 IEEE. Personal use is permitted, but republication/redistribution requires IEEE permission.
See https://www.ieee.org/publications/rights/index.html for more information.}}

\begin{document}

\maketitle

\begin{abstract} 
    In mixed-initiative co-creation tasks, wherein a human and a machine jointly create items, it is important to provide multiple relevant suggestions to the designer. Quality-diversity algorithms are commonly used for this purpose, as they can provide diverse suggestions that represent salient areas of the solution space, showcasing designs with high fitness and wide variety. Because generated suggestions drive the search process, it is important that they provide inspiration, but also stay aligned with the designer’s intentions. Additionally, often many interactions with the system are required before the designer is content with a solution. In this work, we tackle these challenges with an interactive constrained MAP-Elites system that leverages emitters to learn the preferences of the designer and then use them in automated steps. By learning preferences, the generated designs remain aligned with the designer's intent, and by applying automatic steps, we generate more solutions per user interaction, giving a larger number of choices to the designer and thereby speeding up the search. We propose a general framework for preference-learning emitters (PLEs) and apply it to a procedural content generation task in the video game Space Engineers. We built an interactive application for our algorithm and performed a user study with players.
\end{abstract}


\section{Introduction}

    \begin{figure}[!ht]
        \centering
        \includegraphics[width=.48\textwidth]{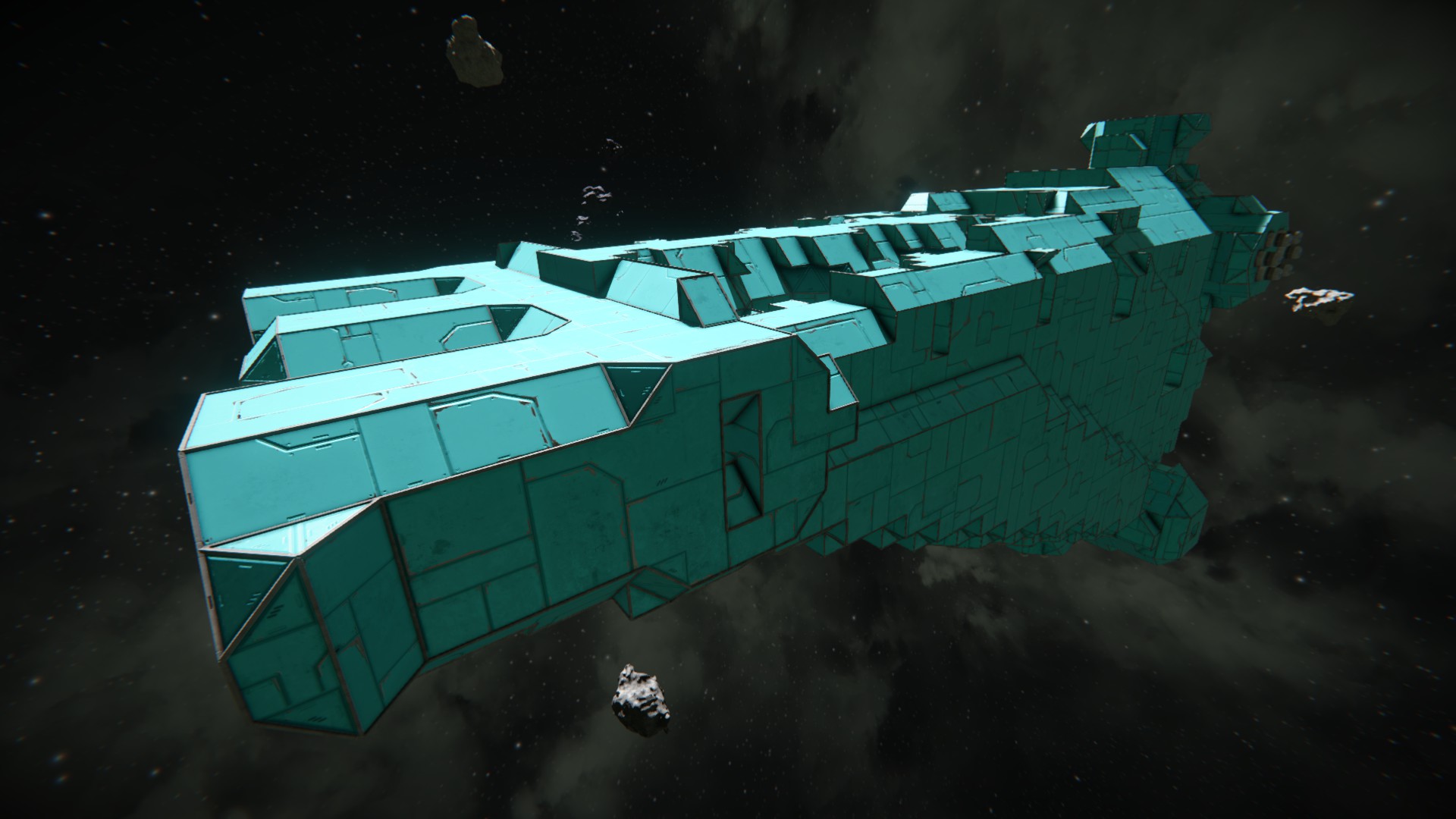}
        \includegraphics[width=.48\textwidth]{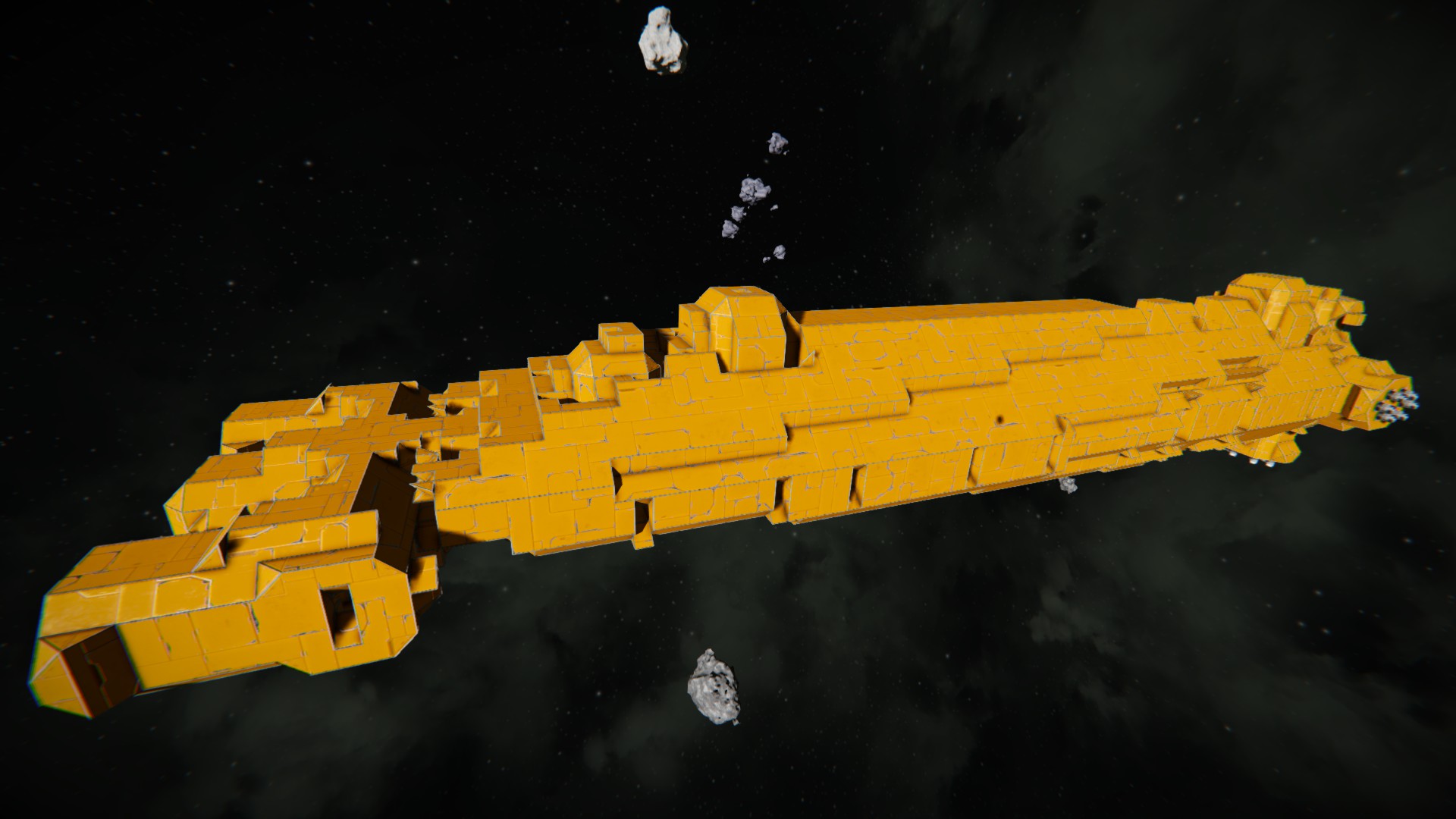}
        \caption{Spaceships generated with our system in the game Space Engineers.}
        \label{fig:user_spaceships}
    \end{figure}
    
    \noindent \IEEEPARstart{A}{s} video games have grown in popularity and size, the problem of automatically generating content for games has become increasingly important. Techniques that aim to solve this problem fall under the umbrella of procedural content generation (PCG) algorithms \cite{shaker_procedural_2016}, which have been used successfully to generate many different types of content, ranging from textures and assets to even entire storylines \cite{hendrikx_procedural_2013}. The most common approaches to PCG use search-based methods, wherein a solution is found by exploring the space of possible solutions using different heuristics. One particularly popular choice in PCG search-based methods are evolutionary algorithms (EAs), which can also have the advantage of finding a set of solutions, instead of just one.
    
    While many PCG algorithms create content autonomously following predefined specifications, there are also methods that continuously leverage a designer's feedback. This paradigm is referred to as mixed-initiative co-creation \cite{yannakakis_mixed-initiative_2014}, and empowers a human designer with the capacity of a computer to generate new suggestions based on human- and/or computer-generated content. Interactive EAs have been used in this setting, where they can propose \enquote{mutated} variations of the current content \cite{yannakakis_mixed-initiative_2014,baldwin_mixed-initiative_2017}.
    
    One family of EAs that have been widely used for PCG \cite{alvarez_empowering_2019,khalifa_talakat_2018,viana_illuminating_2022,zhang_deep_2019} are quality-diversity (QD) algorithms \cite{pugh_quality_2016}, as they provide a set of solutions with both high \enquote{fitness} and diverse characteristics. 
    One classic QD algorithm, the multi-dimensional archive of phenotypic-elites (MAP-Elites) \cite{mouret_illuminating_2015}, has been popular in the mixed-initiative setting \cite{alvarez_fostering_2018,alvarez_assessing_2021}, as it presents solutions in an easy-to-interpret form. This is because it typically projects all solutions onto a 2D grid, ordering solutions along the axes according to different design features called \enquote{behavioural characterisations/descriptors} (BCs). In the mixed-initiative setting, the user can select solutions from this grid for the underlying EA to use as \enquote{parents}, which are then used to create \enquote{offspring} solutions, thereby incorporating human guidance into the search process. At each generation, the number of offspring solutions produced is dependent on the number of parent solutions selected by the human user. As a result, by default, the number of new solutions added to the grid is rather low (because the user will likely only select a few of the presented options), and the difference in fitness between parents and offspring may be small as well. This limitation can be overcome by running additional, automated generations, hidden from the user \cite{alvarez_empowering_2019}.
    
    However, by default in MAP-Elites, and hence in prior work \cite{alvarez_empowering_2019}, the automatic selection of parents is performed at random. We argue that it is better to instead inform the selection via modelling the designer's preferences. To do so, we build upon the \enquote{emitters} framework \cite{fontaine_covariance_2020,cully_multi-emitter_2021}, wherein the selection process in MAP-Elites is instead governed by a learned heuristic. In this work, we introduce a general framework for \enquote{preference-learning emitters} (PLEs) to use in the mixed-initiative setting with MAP-Elites. We test our PLEs in a PCG task of generating spaceships for the game Space Engineers, extending prior work in this domain \cite{gallotta_evolving_2022,gallotta_surrogate_2022} to the mixed-initiative setting. In an internal study, we find that PLEs reduce the amount of time needed to find both playable and visually-appealing spaceships (\Cref{fig:user_spaceships}). We also performed a user study with players of Space Engineers, but were unable to find a statistically significant difference between PLEs and our baselines.

\section{Background}\label{sec:background}

    \subsection{Procedural content generation}\label{subsec:pcg}
        \noindent PCG techniques have been used to create content for many different domains, ranging from robotics to video games \cite{arnold_testing_2013,aycock_procedural_2016}. In the video game creation pipeline, PCG has many benefits: it reduces the workload of designers, it can produce a vast quantity of content in little time, and it can introduce variations within the video game content, which makes the game more interesting for the player.
        
        There are several approaches to PCG, including search-based methods, planning, answer set programming and generative grammars \cite{shaker_procedural_2016}, with search-based methods being the most common.
        
        EAs are a popular choice of search algorithms to use for PCG tasks \cite{gallotta_evolving_2022,gravina_procedural_2019,khalifa_learning_2022,schaa_generating_2021} due to their ability to find collections of solutions. Broadly, EAs are a group of biologically-inspired algorithms in which a \emph{population} (collection) of solutions (each represented by its \emph{genotype}) is created, and each individual is evaluated according to a \emph{fitness function}. A subset of \emph{parent} solutions are picked from the population and new \emph{offspring} solutions are generated from the parents; when performed via genetic operators (such as \emph{crossover} and \emph{mutation}), this subset of EA algorithms are commonly known as genetic algorithms (GAs). The creation of offspring marks the beginning of a new \emph{generation}, and the process is repeated until a termination criterion is met. QD algorithms \cite{pugh_quality_2016} are a family of EAs with the goal of providing a diverse collection of high fitness (\enquote{quality}) solutions for a given task. 
        
        In recent years, QD algorithms have become more widely used for PCG, as they find a broader selection of content compared to other types of EAs. In our domain of interest, which is the creation of spaceships for Space Engineers (\Cref{sec:domain}), prior work used the MAP-Elites QD algorithm to automatically generate a wider variety of spaceships compared to a more standard EA \cite{gallotta_surrogate_2022}.

        An appealing paradigm within PCG is that of mixed-initiative co-creation \cite{yannakakis_mixed-initiative_2014}, in which a human designer can guide the search process interactively. The designer is free to select from different content options proposed by the algorithm, and the algorithm continues the search process from the selected option(s). This approach is less tiring than a completely manual design process, and better able to respect the designer's preferences than a completely automated PCG process. EAs have also been successfully applied in this setting  \cite{alvarez_assessing_2021,baldwin_mixed-initiative_2017,walton_evaluating_2021}. In this work, we extend the hybrid generative-grammar-based EA used for generating Space Engineers spaceships \cite{gallotta_evolving_2022} to the mixed-initiative setting.
    
    \subsection{Quality-Diversity}\label{subsec:qd}
        \noindent Inspired by the idea that evolution is a diversification machine rather than an optimisation system, QD algorithms define \emph{niches} in the search space based on the behavioural characterisations (BCs) of solutions. At the end of the search, each filled niche has at least one solution of high fitness and there is a diversity of solutions as defined by the space over BCs, which are vectors that describe the behaviour of solutions (in biological terms, the \emph{phenotype} of the individual). By keeping \emph{elites}, which are the highest-fitness solutions in different niches, QD algorithms are able to explore the search space more widely than purely fitness-based optimisation algorithms.

        One of the most popular QD algorithms is MAP-Elites \cite{mouret_illuminating_2015}, which projects niches as a 2D grid over BCs. MAP-Elites explores the search space by choosing parents from one of the bins within the grid, creating offspring from these, and then placing the new solutions into their respective bins according to their BCs. In the original MAP-Elites, the selection process of the bins is random: a bin is chosen with uniform probability among all non-empty bins. However, the selection mechanism can be improved to obtain either better fitness, better coverage, or a combination of both. Inspired by the covariance matrix adaptation evolution strategy (CMA-ES), CMA MAP-Elites (CMA-ME) \cite{fontaine_covariance_2020} formalises the concept of emitters, which can provide a more informed selection over bins. Implementing different emitters in MAP-Elites is a promising research direction, as they can significantly alter the overall search process \cite{fontaine_covariance_2020,cully_multi-emitter_2021,gallotta_evolving_2022}.
        
        Another direction of research with QD algorithms is their use in PCG. In order to incorporate design constraints, MAP-Elites has been extended to the constrained optimisation setting via constrained MAP-Elites (CMAP-Elites) \cite{khalifa_talakat_2018}, which uses the feasible-infeasible 2-population (FI-2Pop) constrained optimisation genetic algorithm \cite{kimbrough_feasibleinfeasible_2008} as its base EA. CMAP-Elites has been further extended to the mixed-initiative setting via interactive CMAP-Elites (IC MAP-Elites) \cite{alvarez_interactive_2022}, in which the user guides the selection of parents at every iteration. In this work, we extend prior research \cite{gallotta_surrogate_2022} combining surrogate fitness models, emitters and CMAP-Elites to use IC MAP-Elites for co-creation, and further introduce preference-learning emitters (PLEs) (\Cref{sec:emitters}), which are the main novel contribution of our work.
    
    \subsection{Recommender systems}\label{subsec:recsys}
        \noindent Our PLE framework is inspired by the recommender system problem, in which the goal is to promote items that are relevant to a user by leveraging the relationships that exist between users and items \cite{bobadilla_recommender_2013,jain_trends_2015}. 
        There are many different approaches to constructing recommender systems, with the most common being \emph{content-based filtering}, \emph{model-based collaborative filtering}, and hybrids of these and/or other approaches \cite{aggarwal_recommender_2016, bobadilla_recommender_2013, liphoto_survey_2016}. The term \enquote{hybrid recommender systems} is also used to refer to recommender systems that make use of techniques developed in other areas of artificial intelligence, such as GAs \cite{alcaraz-herrera_evorecsys_2022,ho_hybrid_2007,hwang_using_2010}.
        

        In content-based filtering systems the recommendation is based on finding a relationship between the \emph{content} (properties) of an item and a user profile. In our work, we apply the principles of content-based filtering to find a mapping between the solutions generated by a GA-based PCG algorithm and the user's preferences; thereby we can consider our PLE framework to be a hybrid recommender system.

        One problem with content-based filtering methods is that they can suffer from \emph{overspecialisation}, where the system continuously suggests items that the user has seen previously, and does not suggest any novel items \cite{aggarwal_recommender_2016}. This is related to the \emph{exploration-exploitation dilemma} in reinforcement learning, wherein an agent may start exploiting actions that are known to be rewarding, without exploring alternative, potentially more optimal actions \cite{sutton_reinforcement_2018}. The ability for a recommender system to promote novel and relevant items is known as \emph{serendipity}. One way that recommender systems can achieve such serendipity is by using techniques from the field of multi-armed bandits (MABs) \cite{kohli_fast_2013}, and hence we incorporate MAB sampling strategies (\Cref{subsec:sampling}) within our PLE framework.

        
        Recommender systems can be trained offline and then deployed, or trained online from the start. Both EAs \cite{hinojosa-cardenas_multi-objective_2020,horvath_evolutionary_2017,kim_new_2014,sadeghi_recommender_2017} and deep learning methods \cite{zhang_deep_2019} have been used for offline-trained hybrid recommender systems, whilst MABs \cite{kohli_fast_2013} have been preferred for online training. As we focus on PCG in the mixed-initiative setting, our PLE framework is trained online, and is agnostic to the choice of model for learning user preferences (\Cref{subsec:model}).
        
        

    \subsection{Preference learning}\label{subsec:preflearn}
        \noindent Preference learning is the problem of ordering items according to the preferences of an user \cite{furnkranz_preference_2010}. This can either be expressed as relative rankings (preferring \textit{item A} over \textit{item B}), or via an absolute rating value. The latter can be obtained either explicitly from users, or can be inferred implicitly, e.g., via click counts or number of visits.
        
        
        There are a variety of applications of preference learning in PCG. For example, predicted preferences can be used to alter the fitness in EAs, either by adding a predicted preference value \cite{alvarez_learning_2020}, or by penalising it proportionally to the drift from predicted areas of user interest \cite{hagg_modeling_2019}. Recent work has also investigated the evolution of the design process, clustering offline data and modelling \enquote{designer personas} over time \cite{alvarez_designer_2022}. Nonetheless, one of the main applications is in recommender systems, where the objective is to promote items that the user would be interested in.
        
        In this work, we connect the problem of generating and highlighting relevant solutions in co-creation with the object ranking problem in preference learning, wherein the goal is to learn a ranking function $f(\cdot)$ over a set of items $Z$. We take the user's item selections as implicit ratings, and learn a ranking function that can be applied to all solutions in the population (\Cref{sec:emitters}).
        
    
    \subsection{Multi-armed bandits}\label{subsec:mab}
        \noindent In the MAB setting, there are multiple different actions (\enquote{arms}) that can be taken, each with an associated \emph{reward}, and the goal is for a learner to pick the action with the highest expected reward \cite{jones_dynamic_1972}. As the relationship between actions and their rewards is initially unknown, this gives rise to the exploration-exploitation dilemma. Hence MAB selection strategies focus on finding a balance between exploration and exploitation in order to maximise the reward in the long-term.
        
        One of the simplest MAB selection strategies is $\epsilon$-greedy and its variants \cite{sutton_reinforcement_2018}, wherein with probability $\epsilon$ a random action is chosen, and otherwise the currently predicted optimal action is chosen. There are many other selection strategies, some of which explicitly model a probability distribution over the reward. Commonly-used algorithms of this form include Thompson sampling (TS) \cite{thompson_likelihood_1933} and the upper confidence bound (UCB) \cite{auer_using_2003}. Our PLE framework incorporates MAB selection strategies within the emitter process to balance exploration and exploitation, and hence achieve serendipity (\Cref{subsec:sampling}).
        
        
        MABs can also be modified to the \emph{non-stationary} setting (for example, in recommender systems where users interests may change over time). MAB algorithms designed specifically for this setting include f-discounted-sliding-window Thompson sampling (f-dsw TS) \cite{cavenaghi_non_2021} and sliding-window UCB \cite{wei_nonstationary_2021}. The use of a \enquote{sliding window} is a simple way for us to capture non-stationarity in our PLE framework (\Cref{subsec:selection_history}).
        
        
        MAB algorithms have been applied in many areas within artificial intelligence, including EAs. For example, they have been used as a way to automatically tune hyperparameters in EAs, such as tournament sizes \cite{mouret_quality_2020}, automating selection with a clear exploration-exploitation trade-off \cite{gaier_data-efficient_2017,gaier_discovering_2020,sfikas_monte_2021}, and selecting emitters in QD algorithms \cite{cully_multi-emitter_2021, gallotta_surrogate_2022}. MABs have also been combined with EAs for optimisation within video games \cite{goodman_weighting_2020,liu_rolling_2016}. Our work focuses on the use of MAB algorithms within the emitter process.
        

\section{Preference-Learning Emitters}\label{sec:emitters}
    
    \noindent As introduced in \Cref{subsec:qd}, an emitter drives the selection process of QD algorithms. In the case of MAP-Elites, the emitter operates on the set of solutions in the 2D matrix of bins, which we define as the \emph{container} $Z$. We can also define the subset of occupied bins as $Z_o \subseteq Z$, where a bin is occupied if it has at least one solution from either the feasible or infeasible populations. Each bin can be identified by its $i$ and $j$ indices in the matrix representation of $Z$, which we denote as $b_{i,j}$.
    
    In standard MAP-Elites, the emitter simply selects a random bin $b_{i, j} \in Z_o$ with uniform probability. This type of emitter is therefore called a \textit{random emitter}. Recent work (see \Cref{subsec:qd} for more details) has introduced emitters that select bins to optimise for better fitness, better coverage (defined as the ratio between $|Z_o|$ and $|Z|$), or a combination of both \cite{cully_multi-emitter_2021,gallotta_evolving_2022}.
    
    In IC MAP-Elites, there is no emitter process; instead, the user selects the bin at every iteration.\footnote{For simplicity we restrict the user to select one bin per iteration, but our framework could be expanded to include multiple bin selections.} However, this tightly-coupled interaction slows down the search process. One way to ameliorate this problem is to add extra automated steps, but the random emitter, or even fitness-optimising emitters, are unlikely to match the user's selection criteria. Therefore, in this paper we propose a framework for learning emitters that model the user's preferences, in order to lessen the burden on the user while respecting their intentions.
        
    \begin{figure*}[!t]
        \centering
        \includegraphics[width=\textwidth]{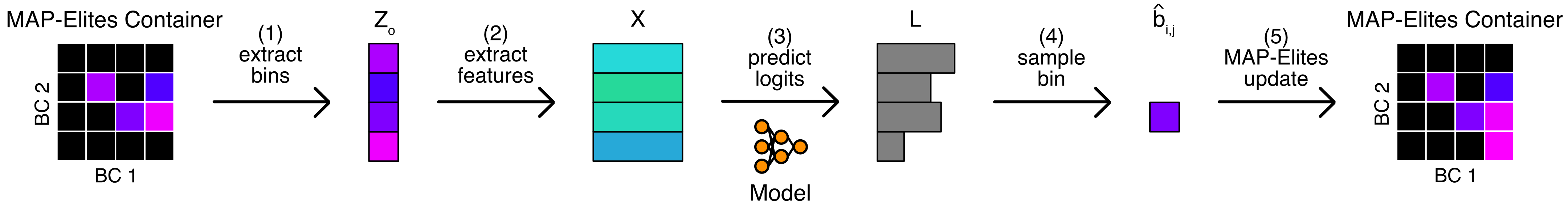}
        \caption{Preference-learning emitter step. (1): The occupied bins, $Z_o$, are extracted from the MAP-Elites container. (2): The input features, $X$, are extracted from $Z_o$. (3): Given $X$, the model predicts $L$, the logits of a categorical distribution over $Z_o$. (4): A bin to use for evolution, $\hat{b}_{i,j}$, is picked from $L$ given the chosen sampling strategy. (5): MAP-Elites performs an update using solutions from $\hat{b}_{i,j}$.}
        \label{fig:ple-diagram}
    \end{figure*}
    
    \begin{table}[!t]
    \centering
    \resizebox{\linewidth}{!}{%
       \begin{tabular}{@{}lc@{}}
            \hline\hline
            Component         & Possible Values                                \\ \hline
            Selection History & $1, k, \dots, \infty$                          \\
            Input Features    & None, Solution (BCs, genotype, phenotype...)   \\
            Model             & Tabular, Non/parametric Non/linear             \\
            Sampling Strategy & Greedy, $\epsilon$-greedy, Boltzmann, Thompson \\ \hline
        \end{tabular}%
        }
    \caption{Components and their possible values in our preference-learning emitter framework. The selection history is a sliding window over user selections (implicit preferences), which can take integer values $k \in [1, \infty]$. This history determines the \enquote{memory} of the emitter (\Cref{subsec:selection_history}). The tabular model does not utilise any input features, but the other models can take in properties of the solutions themselves (\Cref{subsec:in_features}). The model is trained to map between input features and user preferences (\Cref{subsec:model}). Finally, the sampling strategy controls the exploration-exploitation behaviour of the emitter (\Cref{subsec:sampling}).}
    \label{tab:ple_components}
    \end{table}
    
    Our proposed framework for preference-learning emitters (PLEs) leverages (implicit) preference data from each user to predict their current preferences. Our framework is comprised of the following elements:
    \begin{enumerate}
        \item A \textit{history} of past user selections;
        \item A set of \textit{input features} associated with the selections;
        \item A predictive \textit{model} of user preferences; and
        \item A \textit{sampling strategy} that governs the exploration-exploitation ratio over time.
    \end{enumerate}
    We summarise the main components of our proposed framework in \Cref{tab:ple_components}. \Cref{fig:ple-diagram} shows how an automated step is performed using a PLE: the model, fit to user selection data, is used to sample a bin for an additional MAP-Elites update. 
    
    \subsection{Selection history}\label{subsec:selection_history}
        The user's bin selections can be used to implicitly infer their preferences. The bin selection counts can be used as targets within a regression problem, with the values representing an absolute preference value.
        
        While all data could be kept, users' preferences may change over time. One way to account for these changing preferences is to implement a sliding window over the data, retaining only the $k$ most recent selections. At $k = 1$ the resulting PLE would only take into account the last selection, whereas at $k = \infty$ all data becomes available. Using $1 \leq k \leq \infty$ is a simple way of taking into account non-stationarity in user preferences; alternatively, non-stationarity can be handled explicitly by the model.
        
        We can express the user selections as the 3D \enquote{spatiotemporal} tensor $Y$, where each element $Y_{i,j}^{t}$ at iteration $t \in [T - k, T]$ is defined as:
        \begin{equation}
            Y_{i,j}^{t} = \begin{cases}
                1 & \text{if } b_{i,j}^{t} \text{ was selected},\\
                0 & \text{otherwise}
            \end{cases} \quad \forall b_{i,j}^{t} \in Z_o^t,
        \end{equation}
        where $T$ is the current iteration. Retaining the temporal ordering of selections allows us to use this as additional information in the models.

    \subsection{Input features}\label{subsec:in_features}        
        \begin{table}[!h]
            \centering
                \begin{tabular}{@{}cc@{}}
                    \hline\hline
                    Type & Description \\ \hline
                    None & None \\
                    $BC$ & MAP-Elites BCs \\
                    $S$ & \begin{tabular}[c]{@{}c@{}}Genotype and phenotype descriptors (including BCs)\end{tabular}
                    \\ \hline
                \end{tabular}%
            \caption{List of possible input features. The tabular model does not use input features.}
            \label{tab:input-features}
        \end{table}

        Descriptions of the possible input features are given in \Cref{tab:input-features}. The tabular model does not use any input features. $BC$ and $S \supset BC$ are used by the parametric/nonparametric models, and enable them to predict values for novel bins.\footnote{Using more input features increases the risk of learning spurious correlations, particularly when data is limited, hence the choice of $BC \subset S$.}
        
        We can express the input data in the form of a 3D tensor $X$, where each element $X_{i,j}^{t}$, at iteration $t \in [T - k, T]$, is formed as follows:
        \begin{equation}
            X_{i,j}^{t} = \mathrm{extract\_features}(b_{i,j}) \quad \forall b_{i,j} \in Z_o^t,
            \label{eq:input-matrix}
        \end{equation}
        where $\mathrm{extract\_features}$ is a function that extracts the chosen input features from the given bin.

    \subsection{Model}\label{subsec:model}
        The model is used to predict the user's preferences associated with each bin in $Z_o$. Broadly, the models could be trained to predict the normalised selection count, but prior knowledge can be used to adjust this, for example by incorporating decays for non-stationarity. Given $X$, $Y$, and some way of aggregating information temporally, this turns into a standard regression problem.
        
        However, we can utilise more advanced sampling strategies (\Cref{subsec:sampling}) by modelling user preferences probabilistically. As there are a discrete and finite number of bins in $Z_o$, user preferences could be expressed as a categorical distribution over $Z_o$. We can therefore take the raw predictions of the models as the logits $L$ of this distribution. Unfortunately, the size of $Z_o$ changes over time, limiting the amount of models that could be applied to jointly predicting $L$, and so instead we model each bin independently, and later normalise the values if we wish to calculate the categorical distribution.
        
        In our work we used the following models:
        \begin{enumerate}
            \item Tabular: the simplest baseline is a tabular model that simply averages the bin counts $Y$ over time:
            \begin{equation}
                L = \frac{1}{k}\sum_{t=T-k}^{T} Y^t.
                \label{eq:logits_baseline}
            \end{equation}
            
            We also created a more advanced model, which we call the ($\delta$,$\lambda$)-tabular model, that leverages knowledge of the generative process: this model computes the logits $L$ from the bin counts $Y$ for all the timesteps $t$ available in the selection window, setting each entry either due to direct user selection or by selection of a bin containing an offspring solution. If at iteration $t$ the user selects a bin $b_{i,j}$, then the corresponding logit $L_{i, j}$ is increased by $\delta$:
            \begin{equation}
            	L_{i,j} = \delta \cdot \sum_{t=T-k}^{T} Y_{i,j}^t.
            	\label{eq;logits_update1}
            \end{equation}
            Additionally, if the selected bin $b_{i,j}$ contains any solution generated at the previous step by a different bin $b_{m,n}$, we perform credit assignment backwards in time by increasing the value of $L_{m,n}$: 
            \begin{equation}
            	L_{m,n} \leftarrow L_{m,n} + \delta \cdot \sum_{t=T-k+1}^{T} \gamma(b_{m, n}^t, b_{i, j}^{t-1}), 
            	\label{eq;logits_update2}
            \end{equation}
            where $\gamma(b_{m, n}^t, b_{i, j}^{t-1})$ is defined as:
            \begin{equation}
                \gamma(b_{m, n}^t, b_{i, j}^{t-1}) = \frac{1}{n_s^t} \cdot \begin{cases}
                                1 & \text{if $s \in b_{m, n}^{t - 1}$},\\
                                0  & \text{otherwise}
                                \end{cases} \, \forall s | s_o \in b_{i, j}^{t},
            \end{equation}
            where $s$ is a solution contained in the bin, $s_o$ is an offspring of the solution $s$, and $n_s$ is the total number of solutions generated by $b_{m,n}$. 
            
            In general, the computation of the logits at timestep $T$ for a given bin $b_{i, j}$ can be expressed as:
            \begin{equation}
                L_{i, j} = \delta \cdot \left( \sum_{t=T-k}^{T} Y_{i,j}^t (1 - \lambda) + \sum_{t=T-k+1}^{T} \gamma(b_{m, n}^t, b_{i, j}^{t-1}) \right),
                \label{eq:nonparametric-logits}
            \end{equation}
            where $\lambda \in [0, 1]$ is a linear decay factor that can be used to model non-stationarity when set $> 0$.
            One major downside of these tabular models is that they do not place any probability mass on novel bins, unlike the following machine learning models.
            \item Non/parameteric non/linear: standard machine learning algorithms for regression can be used for modelling user preferences. This includes a wide spectrum of methods, with both parameteric and nonparametric models (characterised by a fixed-size vector of parameters $\Theta$ and a variable number of parameters, respectively), and linear and nonlinear models.
            In this work, we experimented with the following common methods:
            \begin{enumerate}
                \item Linear regression: this model computes $L$ via a linear function ${f(\cdot; \theta): \mathbb{R}^n \rightarrow \mathbb{R}_+}$, where $n$ is the dimensionality of the input features. The model is trained to predict $Y$ averaged over $t$ (equivalent to \Cref{eq:logits_baseline}) using the mean squared error (MSE) loss, and uses the closed-form solution to find the optimal $\theta$. Individual predictions can be expressed as follows:
                \begin{equation}
                    L_{i,j} = f(X_{i,j}^{t}; \theta);
                    \label{eq:parametric-logits}
                \end{equation}
                \item Ridge regression: this extends linear regression to incorporate regularisation on the L2-norm of the parameters;
                \item Neural network (regression): this operates similarly to the linear models, but is able to model nonlinear functions using an artificial neural network. As there is no closed-form solution, the parameters are updated using stochastic gradient descent, starting at the previous weights at each iteration;
                \item $k$-nearest neighbours (kNN) regression: this nonparametric model assigns the logits of novel bins based on their distance from past bins whose bin count is known via the kNN algorithm with distance-based weighting; and
                \item Kernel ridge regression (KRR): this nonparametric model leverages either a linear or nonlinear kernel (such as the radial basis function) to estimate the logits from the bin counts, using the MSE loss with L2-norm regularisation.
            \end{enumerate}
        \end{enumerate}
        
    \subsection{Sampling strategies}\label{subsec:sampling}
        Once we have computed $L$, we can use it to sample a bin from $Z_o$. The most naive method is greedy sampling, which always picks the bin with the highest probability:
        \begin{equation}
        \hat{b}_{i,j} = \argmax_{i,j} L,
        \end{equation}
        where we use $\hat{b}_{i,j}$ to differentiate automated bin selections from human bin selections.
        
        A commonly-used variation on this, $\epsilon$-greedy, can be used as follows:
        \begin{equation}
        	\hat{b}_{i,j} = \begin{cases}
        		\argmax_{i,j} L & \text{with probability } 1 - \epsilon,\\
        		\mathcal{U}(Z_o) & \text{with probability } \epsilon,
        	\end{cases}
        	\label{eq:epsgreedy_sampling}
        \end{equation}
        where either the bin with highest probability is picked, or otherwise a random bin is chosen uniformly from $Z_o$. In practice this sampling strategy is often combined with a decay on $\epsilon$, controlled by a hyperparameter $\lambda$. We use the power law decay:
        \begin{equation}
        	\epsilon \leftarrow \epsilon - \lambda\epsilon;
        	\label{eqn:decay}
        \end{equation}
        ensuring that a variety of solutions are explored initially, but as more data becomes available the selection \enquote{exploits} the best bin.
        
        We can also sample directly in proportion to the probabilities of the categorical distribution:
        \begin{equation}
            \hat{b}_{i,j} \sim \frac{e^{L_{i,j}/\tau}}{\sum_{m,n} e^{L_{m,n}/\tau}} \quad \text{for }m = 1 \ldots i, n = 1 \ldots j,
            \label{eq:boltzmann_sampling}
        \end{equation}
        where $\tau$ is the temperature of the distribution: high temperatures result in a more uniform distribution, whilst low temperatures accentuate high probability elements. This is known variously in the MAB literature as \emph{softmax}, \emph{Boltzmann} or \emph{Gibbs sampling}. We note that because this method selects bins proportionally to their predicted preference value, bins that are predicted to be \enquote{suboptimal} can still be sampled with a small probability. Similarly to $\epsilon$-greedy, the temperature can be decayed over time (\Cref{eqn:decay}) to change the ratio between exploration and exploitation.
        
        Finally, we consider a Bayesian approach to sampling: Thompson sampling (TS). TS places a distribution over the parameters of the preference model, sampling parameters from the posterior distribution as part of the selection procedure.
        
        The simplest tabular model that averages bin counts can be interpreted as predicting preferences with a Bernoulli distribution (where $L_{i,j}$ is the probability $p$ of the distribution). We can therefore use the Beta distribution, parameterised by $\alpha$ and $\beta$, as the conjugate prior over $p$. Given prior values for $\alpha$ and $\beta$, TS with this tabular model proceeds as follows:
        \begin{align}
            L_{i,j} \sim Beta(\alpha_{i,j}, \beta_{i,j}) \\
            \hat{b}_{i,j} = \argmax_{i,j} L.
        \end{align}
        Whenever a bin is picked by the human, the posterior update is as follows:
        \begin{equation}
            \begin{alignedat}{2}
                 &\alpha_{i,j}, \beta_{i,j} \leftarrow \alpha_{i,j} + 1, \beta_{i,j} + 1 &&\text{ if }b_{i,j}\text{ was selected,}\\
                 &\alpha_{i,j}, \beta_{i,j} \leftarrow \alpha_{i,j}, \beta_{i,j} + 1 &&\text{ otherwise}.
            \end{alignedat} 
        \end{equation}
        Unlike Boltzmann sampling, TS explicitly takes into account uncertainty over the predicted values, using this to automatically control exploration vs. exploitation.
    
    \subsection{Process overview}
        
        \RestyleAlgo{ruled}
        \begin{algorithm}[h!]
            \DontPrintSemicolon
            \KwData{MAPElites, Emitter, $n_{steps}$}
            $Z_o \gets \mathrm{extract\_bins}(\text{MAPElites})$\;
            \text{// User step}\;
            $b_{i,j} \leftarrow \mathrm{human\_selection}(Z_o)$\;
            MAPElites.$\mathrm{update}(b_{i,j})$\;
            Emitter.$\mathrm{update}(b_{i,j})$\;
            \text{// Emitter steps}\;
            \For{$n \in n_{steps}$} {
                $Z_o \gets \mathrm{extract\_bins}(\text{MAPElites})$\;
                $X \gets \mathrm{extract\_features}(Z_o)$\;
                $L \gets \text{Emitter}.\mathrm{model}(X)$\;
                $\hat{b}_{i,j} \gets \text{Emitter}.\mathrm{sample}(L)$\;
                $\text{MAPElites}.\mathrm{update}(\hat{b}_{i,j})$\;
            }
        \caption{One iteration of IC MAP-Elites with a PLE}
        \label{alg:emitters_overview}
        \end{algorithm}

        \noindent We now give an overview of an iteration of IC MAP-Elites with a PLE (with pseudo-code in \Cref{alg:emitters_overview}). First, the human user selects a bin $b_{i,j}$, which is used by the underlying EA to generate new solutions. The selection is also used to update the PLE's internal state. Then the PLE is queried for a set number of steps. During each step (previously shown diagramatically in \Cref{fig:ple-diagram}), we extract the features $X$ from the occupied bins in $Z_o$, use the model to generate the logits $L$, sample a bin $\hat{b}_{i,j}$ using the chosen sampling strategy, and finally run another MAP-Elites update.

\section{Domain}\label{sec:domain}
\begin{figure*}
        \centering
        \includegraphics[width=0.33\textwidth]{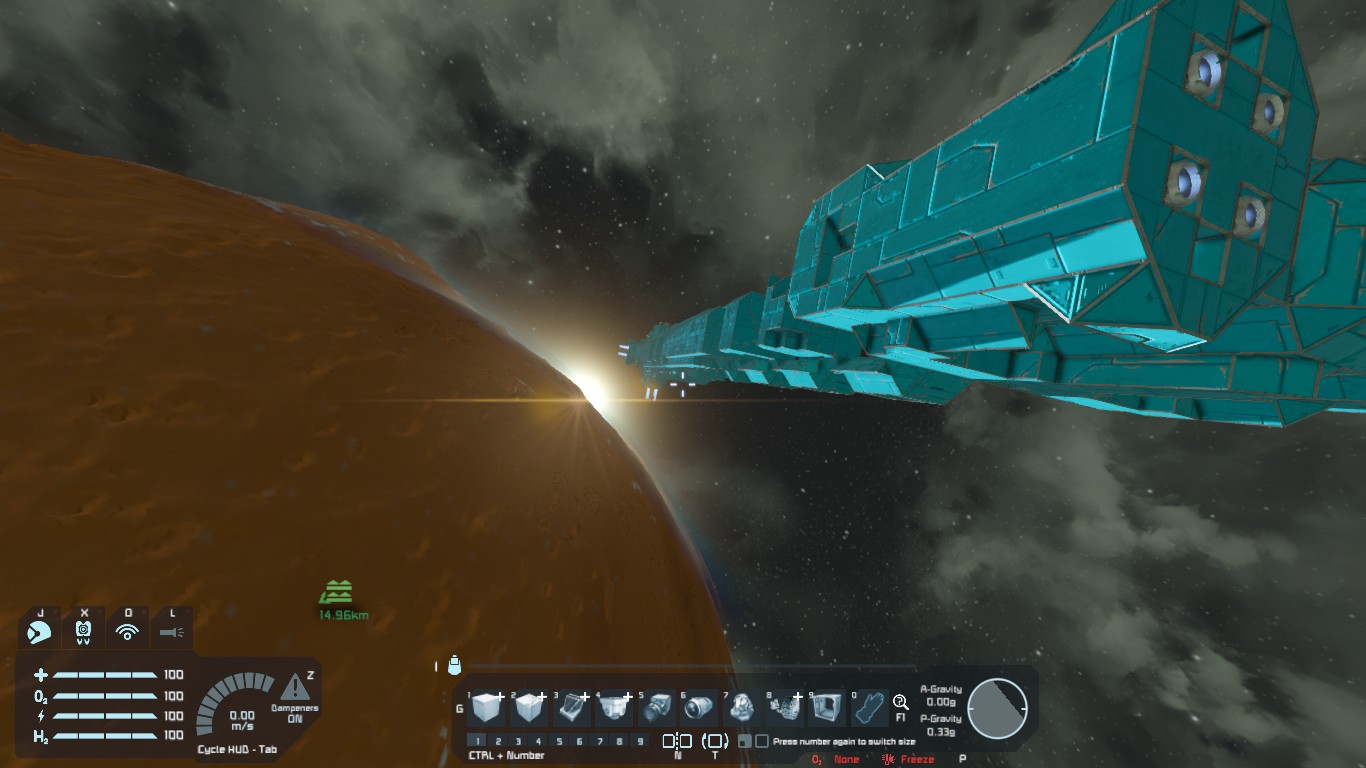}%
        \includegraphics[width=0.33\textwidth]{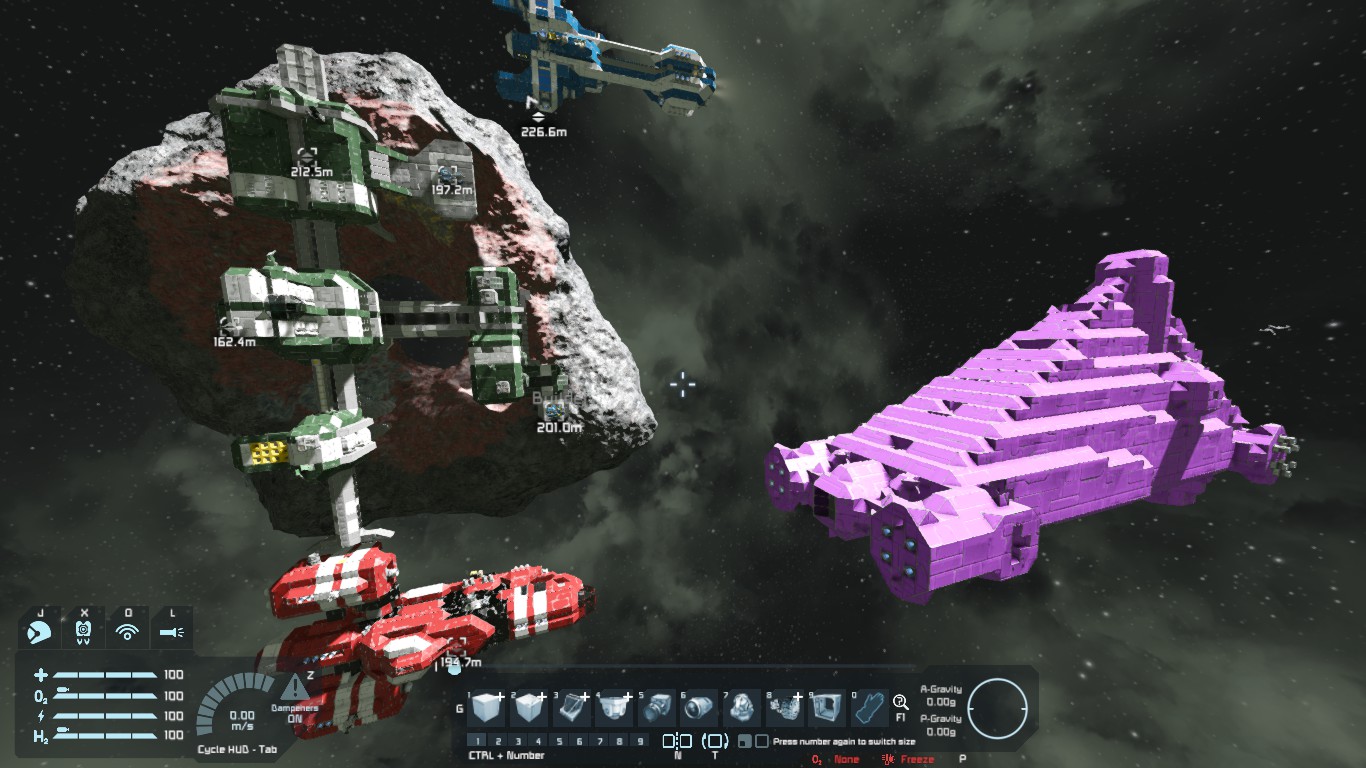}%
        \includegraphics[width=0.33\textwidth]{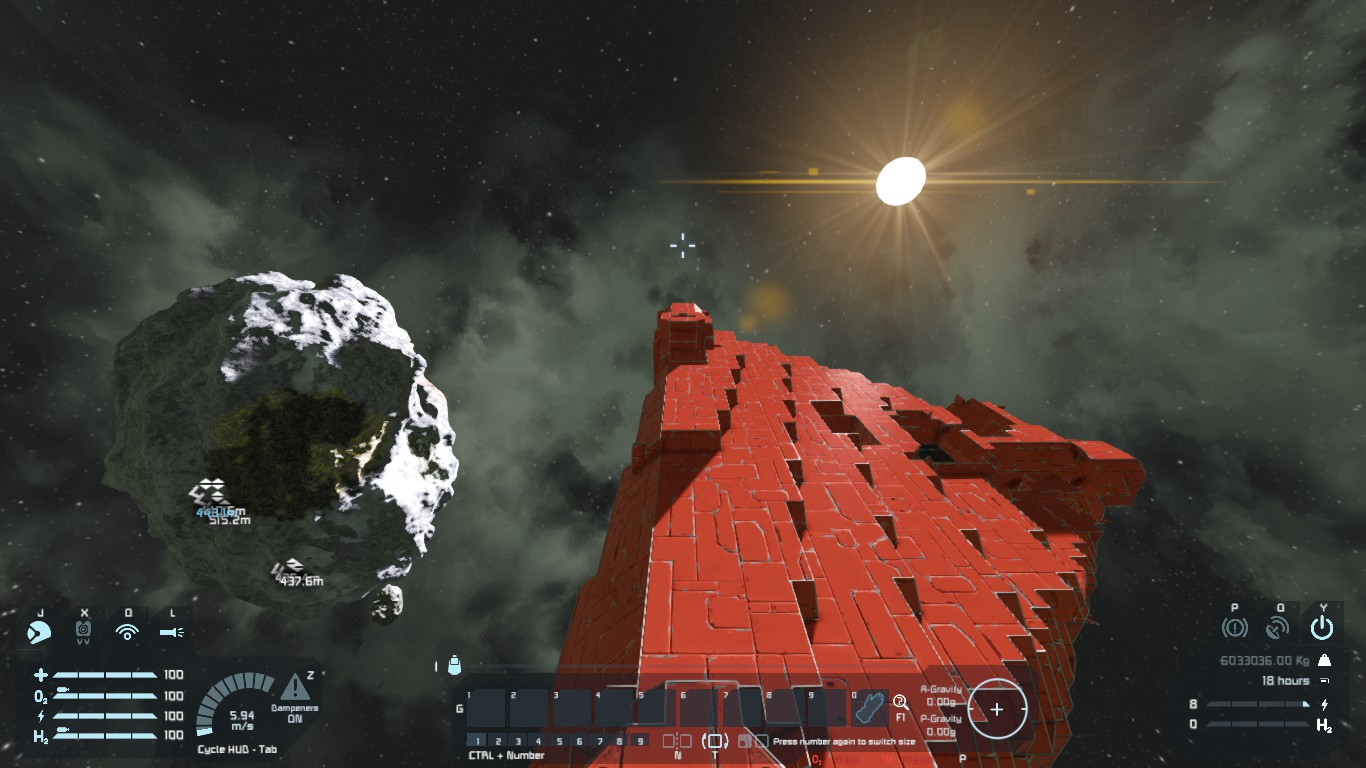}\\
        \includegraphics[width=0.33\textwidth]{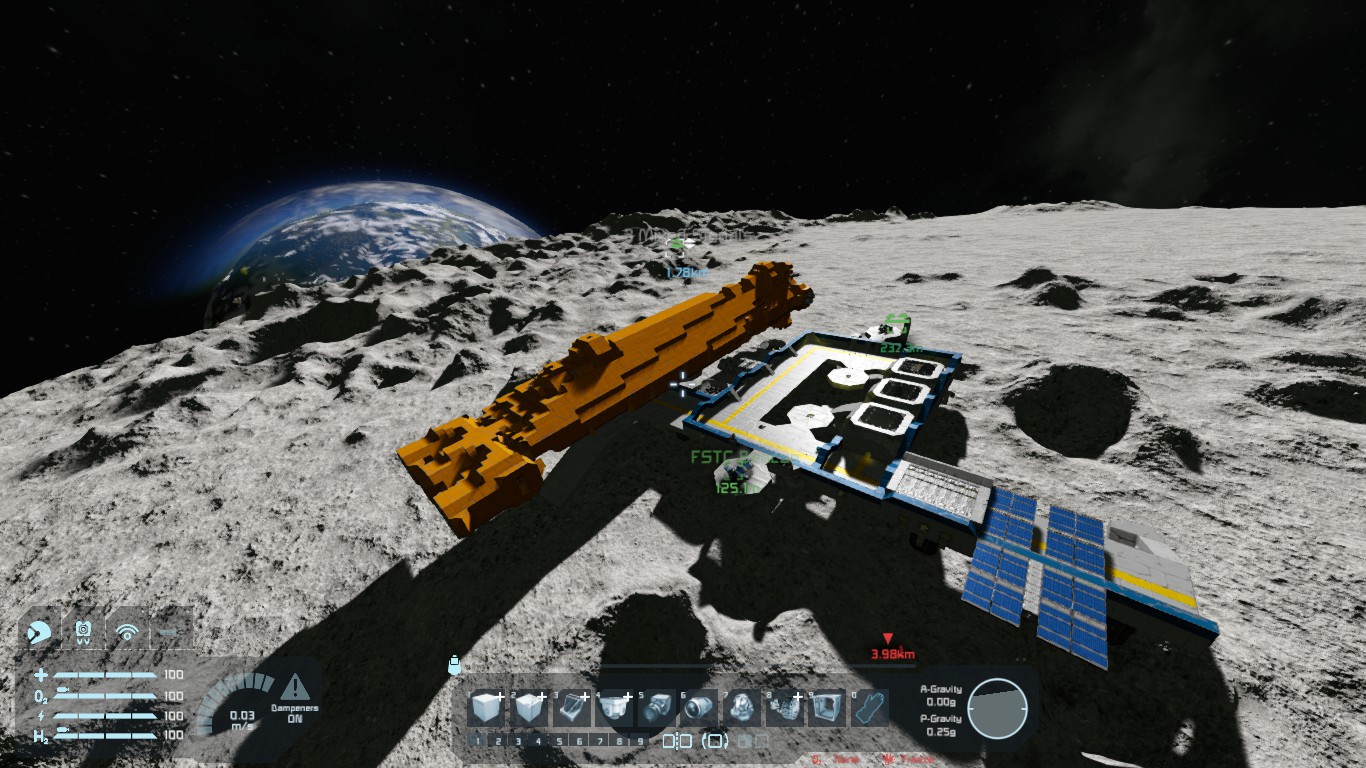}%
        \includegraphics[width=0.33\textwidth]{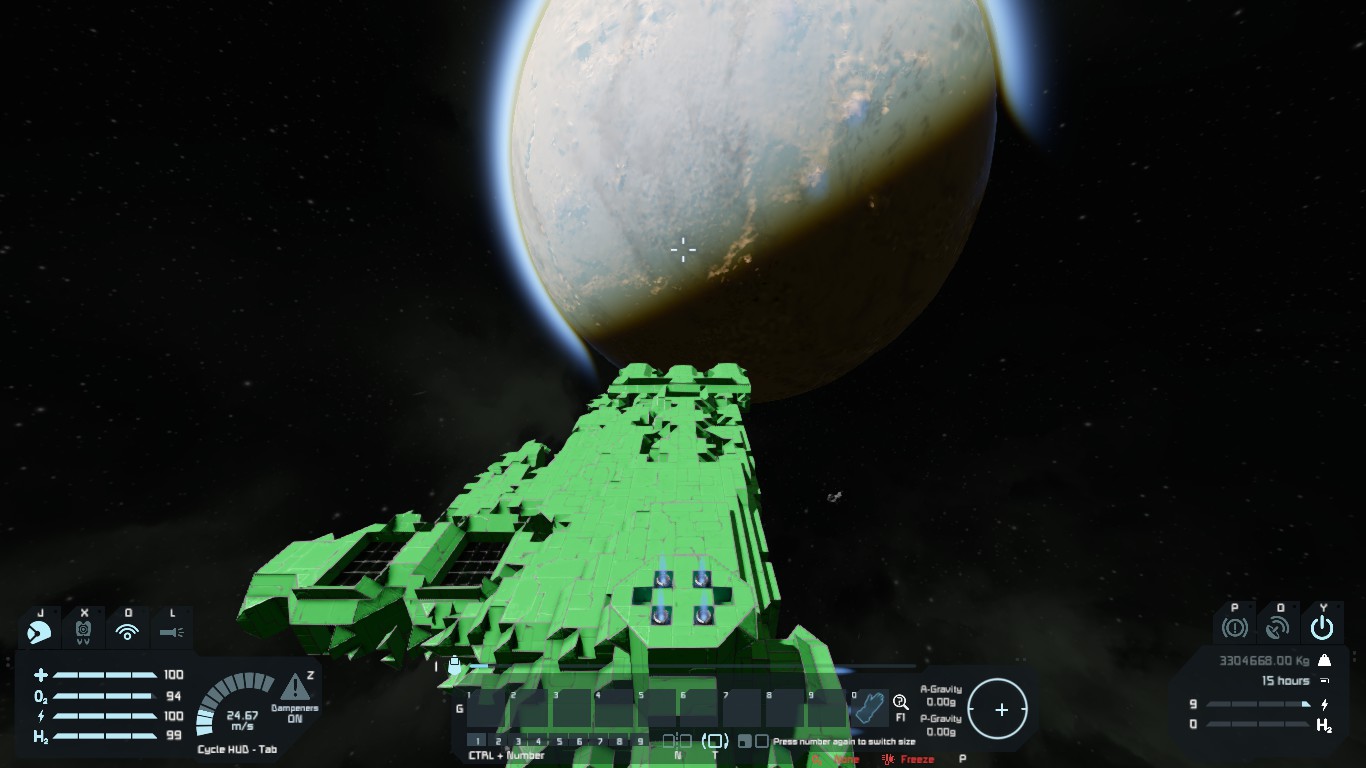}%
        \includegraphics[width=0.33\textwidth]{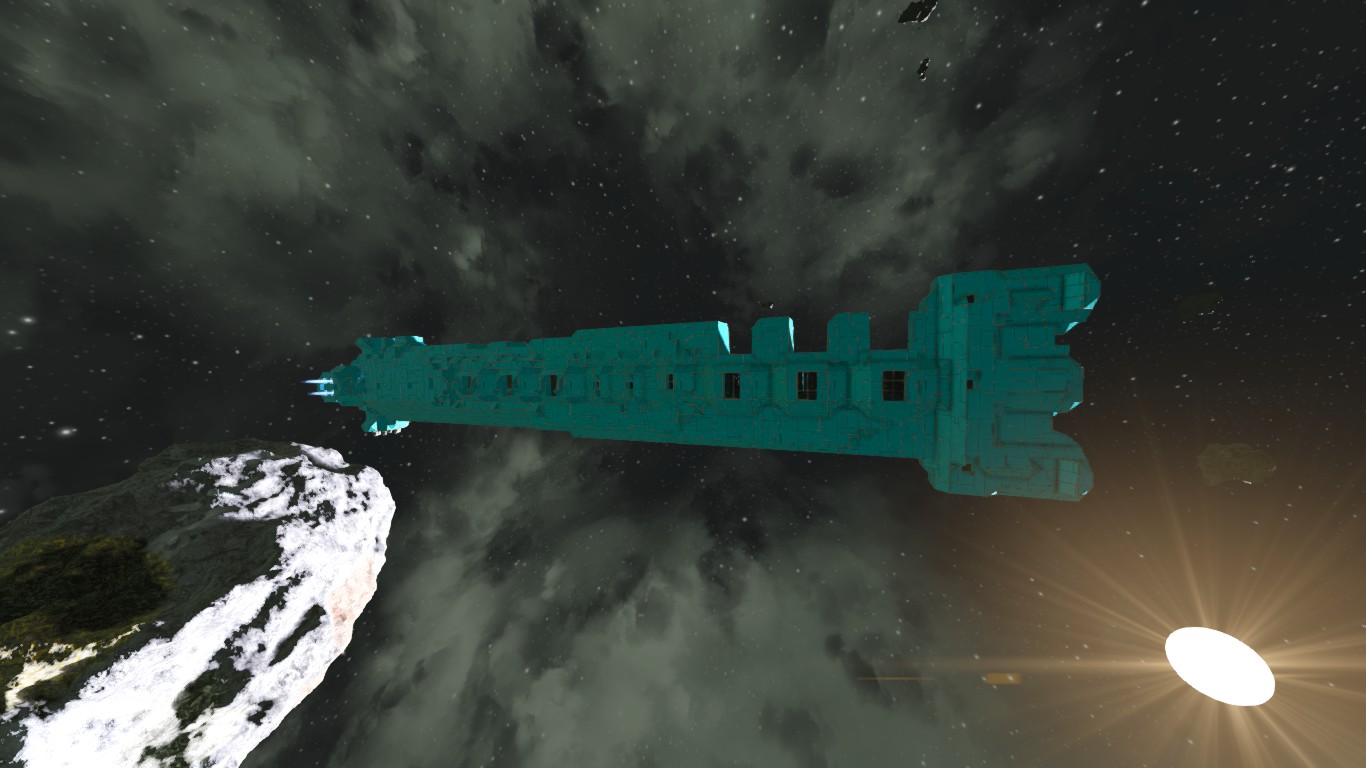}\\
        \caption{Screenshots of spaceships generated using our system in different scenarios in Space Engineers.}
        \label{fig:ingame-spaceships}
    \end{figure*}

    \noindent Space Engineers is a popular 3D sandbox video game with over 5000 daily active users. The game is set in outer space, where the player builds structures and spaceships to mine resources, travel between planets, and fend off enemies. The game fosters creativity by allowing players to freely build block-based structures, enforcing only the necessary restrictions for functionality. One of the primary features of the game is a realistic physics engine, which requires players to consider physical properties when constructing buildings or vehicles. The system developed in this work is capable of generating a variety of spaceships that can be successfully piloted in-game (\Cref{fig:ingame-spaceships}).

    Initial work in this domain \cite{gallotta_evolving_2022} introduced a hybrid EA, combining L-systems \cite{lindenmayer_mathematical_1968} with FI-2Pop to generate spaceships with functional constraints (each spaceship must have the required components in order to operate, e.g., reactors, thrusters, and blocks should not intersect). The genotype is a string consisting of L-system atoms, and the phenotype is the voxel representation of the spaceship. In order to construct a fitness function to emulate human aesthetics, over 200 user-generated spaceships were downloaded from Steam Workshop and used to construct a distribution over different ship properties (phenotype descriptors). The four properties calculated include the following ratios: the amount of functional blocks to the total number of blocks; the filled volume to the total (bounding box) volume; the major axis to the medium axis; and the major axis to the smallest axis. Density models were used to approximate the empirical distribution of these four properties, with the fitness function being defined as the sum of the probabilities of each property under the density models. This procedure and the formulation of the fitness function can be found in more detail in \cite{gallotta_evolving_2022}.
    
    Subsequent work \cite{gallotta_surrogate_2022} then improved upon this process with a novel variant of FI-2Pop, and further extended the hybrid EA to be used within CMAP-Elites, using the two spaceship axis ratios as BCs. In \cite{gallotta_surrogate_2022}, the CMAP-Elites grid was fixed at 32 $\times$ 32 in order to report a coverage metric, but in this work we start at a more human-friendly 10 $\times$ 10 and subdivide a bin into four quadrants whenever it contains 5 solutions, up to a maximum of 4 subdivisions per original bin.
    
    In the current work we shift from fully-automated PCG to the mixed-initiative setting, learning to interactively generate content tailored for different users.

    \begin{figure}[!ht]
        \centering
        \includegraphics[width=.24\textwidth]{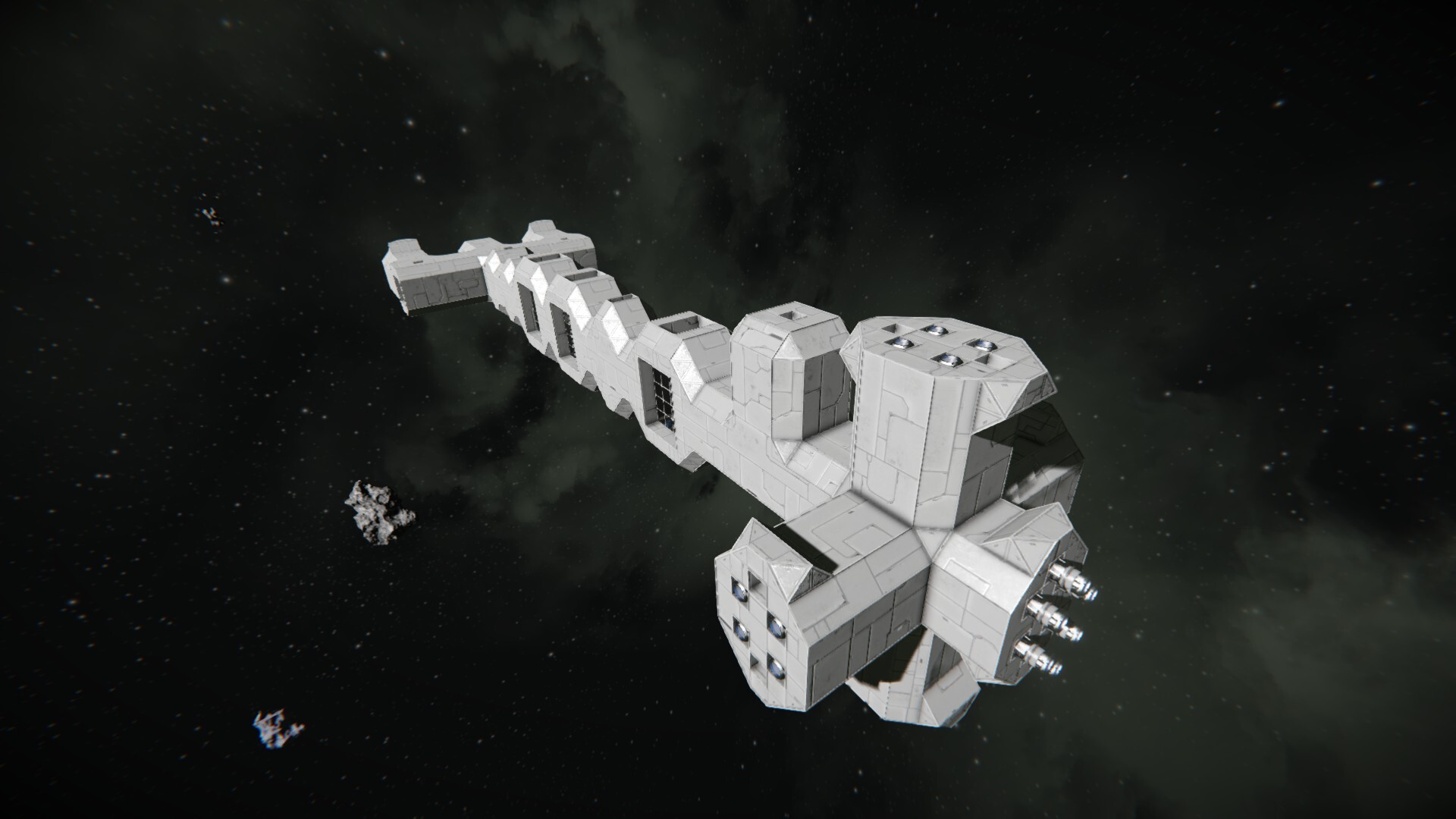}%
        \includegraphics[width=.24\textwidth]{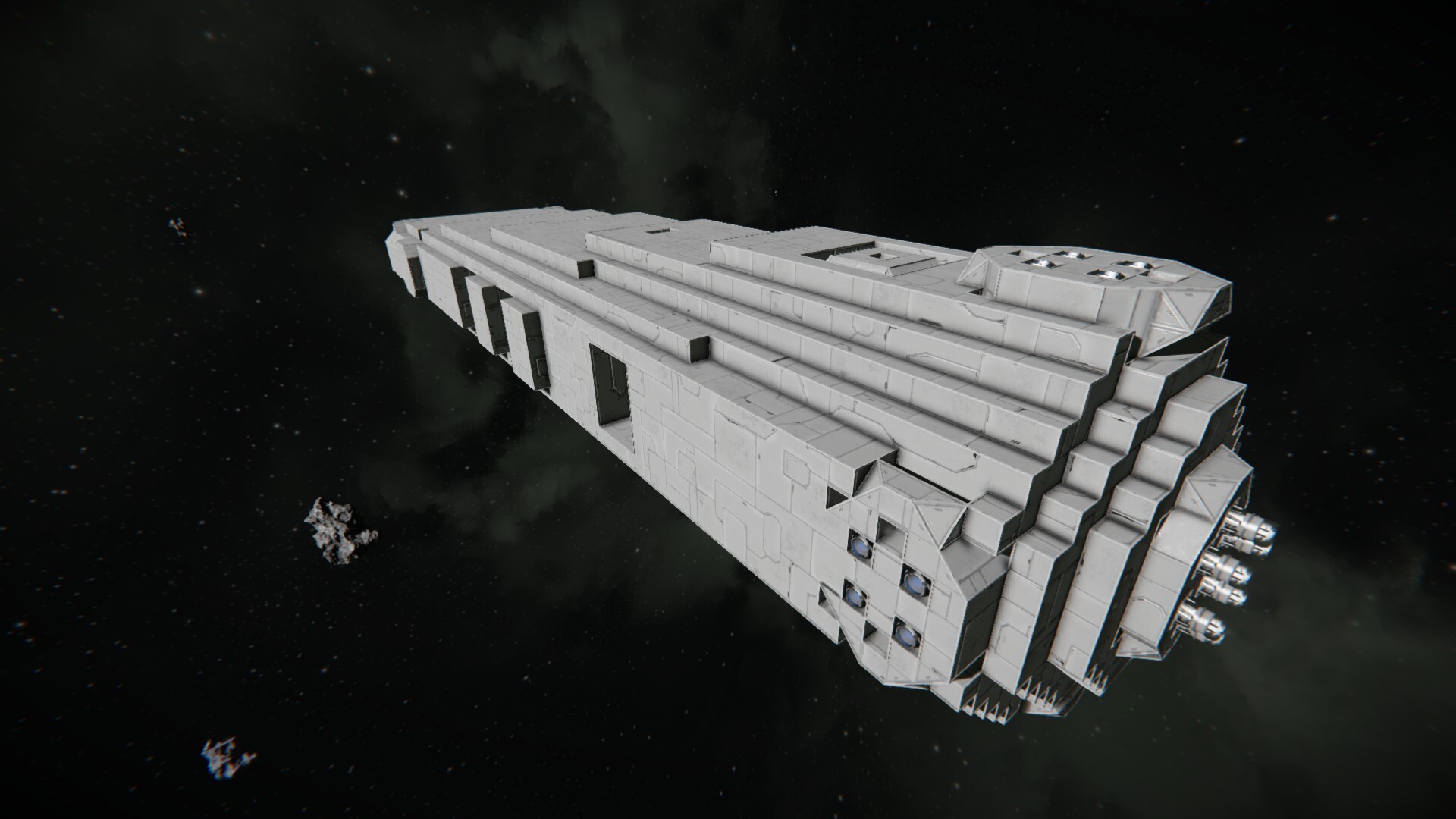}
        \\
        \includegraphics[width=.24\textwidth]{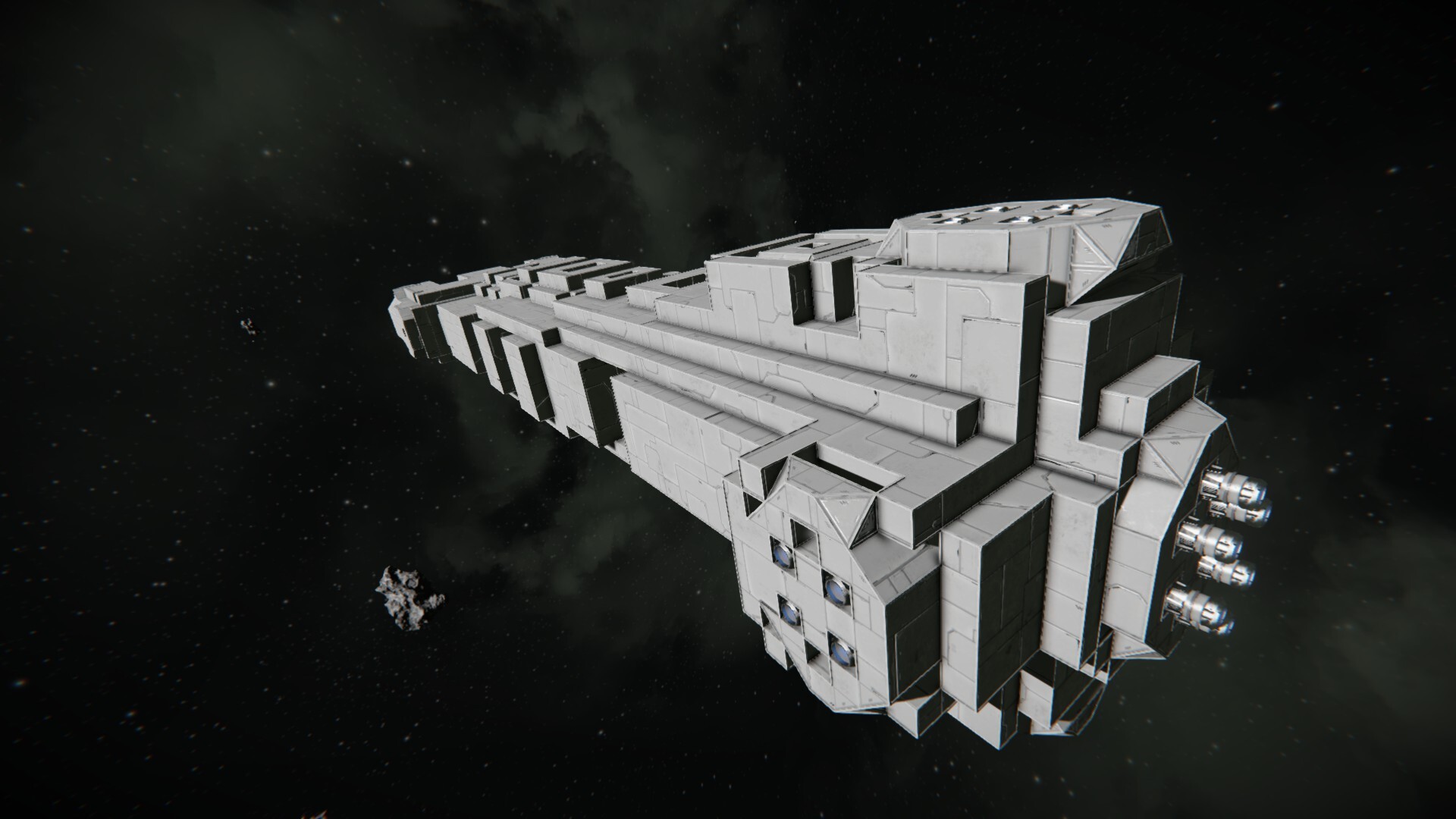}%
        \includegraphics[width=.24\textwidth]{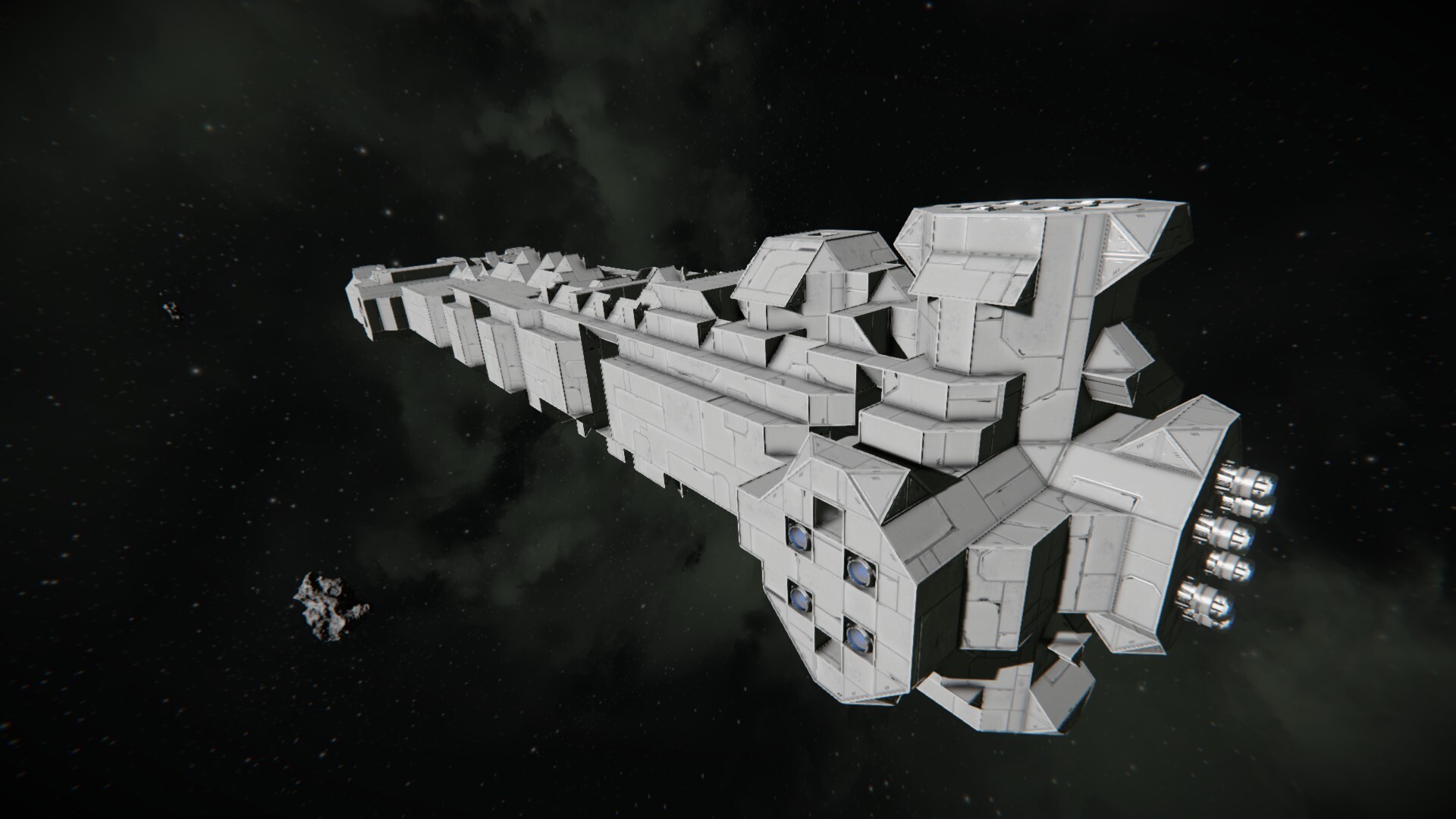}
        \caption{Effects of the different steps of applying the hull-building process to a spaceship. Starting from the initially-generated structure (top left), we first create a convex hull (top right), apply binary erosion (bottom left), and then iteratively smooth it out (bottom right).}
        \label{fig:hullbuilding-steps}
    \end{figure}
    
    One shortcoming of the initial hybrid EA proposed in \cite{gallotta_evolving_2022} was that the spaceships were generated around corridor structures, ensuring that all areas within the spaceship were linked, but resulting in unnatural appearances. In this work, we resolve this issue by adding external hulls to the generated spaceships. To do so, we first create a convex hull \cite{barber_quickhull_1996} comprised of base blocks, and then apply binary erosion \cite{serra_image_1982} to obtain a more organic-looking hull. Finally, we iteratively replace the blocks in the hull with sloped blocks of various types to smooth it out. The entire hull-building process is presented in \Cref{fig:hullbuilding-steps}.
    
    While creating the convex hull and applying erosion is computationally efficient, the iterative smoothing process can become quite expensive for larger spaceships, and so we only apply it when the user decides to download the content from our application. This choice does affect the fitness value of the spaceship; however, the difference is minimal and can be overlooked for the sake of computation time saved.
    
    Finally, a small change was made to the L-system rules to ensure thrusters are placed along all 6 axes. 
    This change gives finer control over acceleration/deceleration, and enables players to dock their ship, but can also be toggled off in the application to explore a larger variety of architectures.

\section{Application}\label{sec:webapp}

    \begin{figure*}[!t]
        \centering
        \includegraphics[width=1\textwidth]{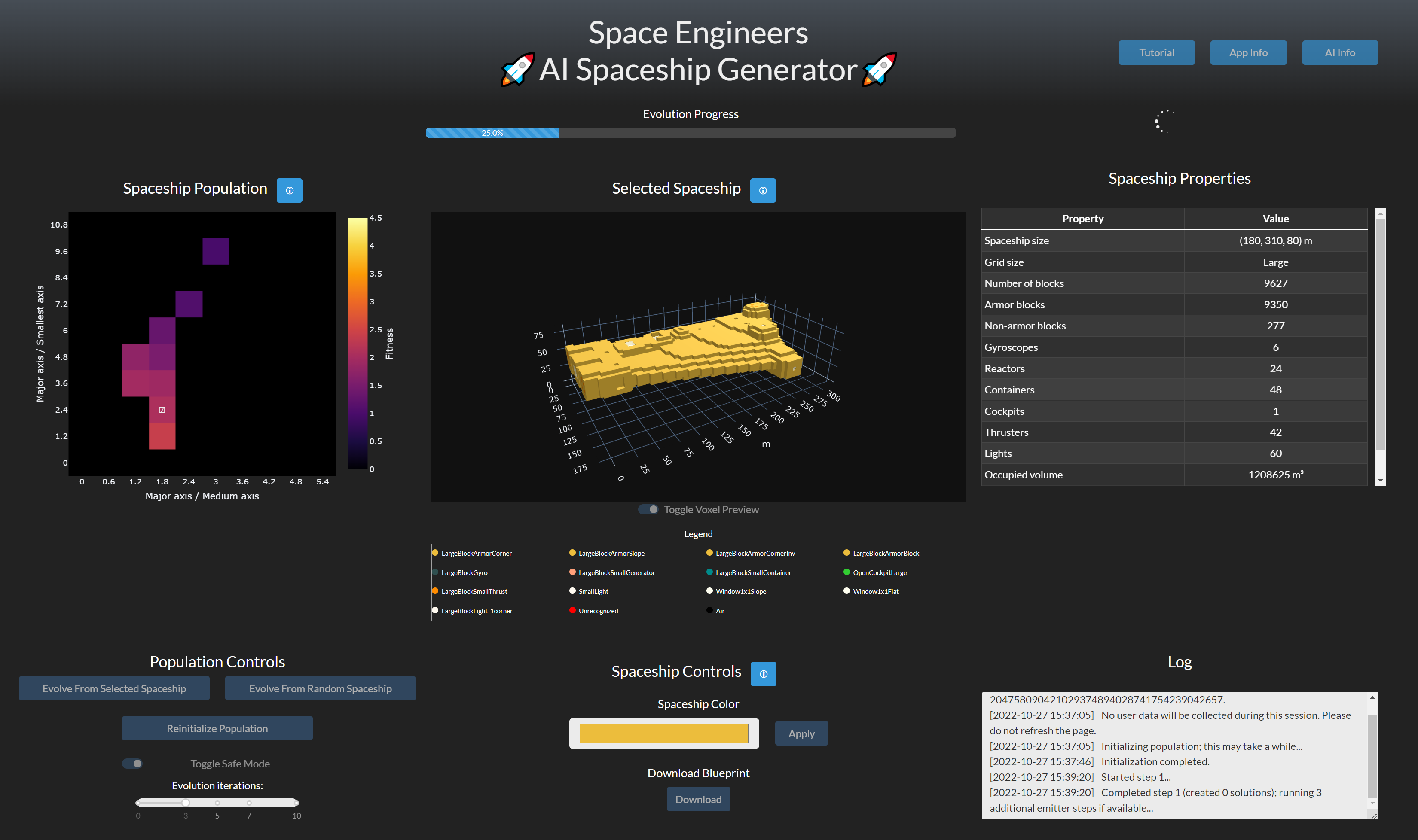}
        \caption{Screenshot of the AI Spaceship Generator application in user mode mid-generation. In the application the user can view the population of spaceships, inspect spaceships and their properties in further detail, generate new spaceships, and download a spaceship blueprint to import into Space Engineers.}
        \label{fig:webapp-user}
    \end{figure*}

    \noindent We built a graphical user interface for our system (\Cref{fig:webapp-user}) using the Dash\footnote{Available at \href{https://dash.plotly.com/}{https://dash.plotly.com/}.} library \cite{hossain_visualization_2019} for creating web applications, and further provide Windows executables, compiled using PyInstaller\footnote{Available at \href{https://github.com/pyinstaller/pyinstaller}{https://github.com/pyinstaller/pyinstaller}}. All code and the applications are available at \href{https://github.com/arayabrain/space-engineers-ai-spaceship-generator}{https://github.com/arayabrain/space-engineers-ai-spaceship-generator}.

    There are six main components in the AI Spaceship Generator application:
    \begin{enumerate}
        \item Spaceship Population: here the user can view the current MAP-Elites grid and select bins on the grid to inspect the elite contained within;
        \item Selected Spaceship: here an interactive 3D preview of the selected elite is shown, with a toggle to see the insides of the spaceship;
        \item Spaceship Properties: here relevant in-game properties of the selected elite are listed;
        \item Population Controls: here the user can apply an evolutionary step or reset the population;
        \item Spaceship Controls: here the user can change the main colour of the spaceship, as well as download a spaceship blueprint file that can be directly imported into the game; and 
        \item Log: here all application messages are displayed to the user.
    \end{enumerate}
    The application also includes a tutorial, a help menu, and information about the underlying generative algorithms.
    
    The application has three different modes available: a \emph{user} mode, a \emph{developer} mode and a \emph{user-study} mode. These modes differ in the amount of control over the system available to the user.
    
    In user mode (\Cref{fig:webapp-user}), the user can interact with the MAP-Elites grid (displaying the feasible population), preview the selected bin's elite spaceship, and inspect its properties. By clicking the \enquote{Evolve from Selected Spaceship} button, a single step of FI-2Pop is applied with the currently selected bin used for parents, and then multiple automated steps are applied using the currently set emitter (by default, a PLE). By clicking the \enquote{Evolve from Random Spaceship} button, the \emph{random emitter} is used to sample the bin to be used in the initial FI-2Pop step. By clicking the \enquote{Reinitialise Population} button, the user is able to reset the current MAP-Elites population. By default, the application is set to \enquote{Safe Mode}, where all spaceships have at least one thruster on all six sides for maximising manoeuvrability; but this can be disabled to create a larger variety of spaceships. Finally, the \enquote{Evolution Iterations} slider allows the user to alter the number of emitter steps: this increases the expected number of solutions generated, at the cost of additional time per user step.
    
    After the initial population is generated and displayed on the MAP-Elites grid, the process to create new spaceships proceeds as follows:
    \begin{enumerate}
        \item the user selects an occupied bin in the \enquote{Spaceship Population};
        \begin{enumerate}
            \item the user clicks the \enquote{Evolve from Selected Spaceship} button, which triggers one iteration of our FI-2Pop variant \cite{gallotta_surrogate_2022} with the currently selected bin, and then multiple automated iterations with the current emitter; or
            \item the user clicks the \enquote{Evolve from Random Spaceship} button which triggers multiple automated iterations with the random emitter;
        \end{enumerate}
        \item the \enquote{Spaceship Population} is updated, and the process can repeat.
    \end{enumerate}
    
    In developer mode, control over the MAP-Elites grid is more advanced, allowing the user to display either the feasible or the infeasible population, select among different metrics (fitness, bin coverage, and age), and view either the elite or the bin's average value of the current metric. More controls over the generative process are also provided, such as changing the active emitter, toggling mutability of L-system modules, changing the MAP-Elites BCs, and changing weights over components in the fitness function. Additionally, in developer mode it is possible to change the L-system rules during the evolution process, allowing for a higher level of control over the content generated.

    In order to test our PLE framework with Space Engineers players (see \Cref{sec:userstudy}), we also designed a user-study mode. In this mode, the user is set up with 1 of 4 different emitters (the choice of which being hidden from the user), and asked to run 6 iterations of evolution, after which the user is asked to select their favourite content of the current generation, which is then downloaded for the final step of the user study. This process is repeated until all emitter configurations have been tested, at which point the application switches to the normal user mode. In user-study mode, the current progress is indicated to users by additional progress bars at the top of the application. To ensure that the emitters are tested under the same settings, the Population Controls are restricted to \enquote{Evolve from Selected Spaceship}. This mode also tracks statistics about the generative process (\Cref{sec:userstudy}).
    
    The final stage of the user study asks users to rank the spaceships they selected in the previous step of the experiment. By uploading the files generated prior to a standalone Spaceships Ranker application, the user is shown the generated spaceships and tasked with ranking them relatively to the others. Once the ranking has been performed, the user can download the result of their ranking to upload on the user study questionnaire.
    

\section{Internal study}\label{sec:prelim_study}

    \noindent We first performed an internal study where we tested different configurations of our generative system based on our PLE framework. We used the following models and associated training settings, as implemented in the scikit-learn library \cite{scikit-learn}:
    \begin{itemize}
        \item Linear: \textbf{linear\_model.LinearRegression};
        \item Ridge: \textbf{linear\_model.Ridge};
        \item Neural network: \textbf{neural\_network.MLPRegressor} with 2 hidden layers, L2 regularisation, and learning rate $\eta = 0.001$;
        \item kNN: \textbf{neighbors.KNeighborsRegressor} with $k = 5$ neighbours, distance-based weights with Euclidean distance metric, and leaf size $= 30$ for approximate kNN; and
        \item KRR: \textbf{kernel\_ridge.KernelRidge} with either a \enquote{linear} or \enquote{rbf} kernel.
    \end{itemize}
    
    With these models we tested the following hyperparameters and associated values (with the best values from our internal study highlighted in \textbf{bold}):
    \begin{enumerate}
        \item selection window value $k$: 2, 5, $\pmb{\infty}$;
        \item solution context $S$: fitness component values and axis sizes, only fitness component values, \textbf{only axis sizes};
        \item increment $\delta$ and decay $\lambda$ values for the tabular model: \textbf{(1, 0.5)}, (1, 0.75), (1, 1);
        \item ridge L2 regularisation term: \textbf{1}, 1e-2, 1e-3;
        \item ridge solver: SVD, Cholesky, \textbf{SGD}, L-BFGS-B;
        \item neural network hidden layer size: (100, 100), \textbf{(200, 200)};
        \item neural network activation function: \textbf{ReLU}, tanh;
        \item neural network optimiser: SGD, L-BFGS, \textbf{Adam};
        \item neural network L2 regularisation term: 1e-4, \textbf{1e-3};
        \item training epochs for the parametric models: 10, \textbf{20}, 50;
        \item kernel ridge L2 regularisation term: \textbf{1}, 1e-2, 1e-3;
        \item initial $\epsilon$ and decay $\lambda$ values for $\epsilon$-greedy sampling (0.2, 0.01), \textbf{(0.9, 0.1)};
        \item temperature $\tau$ and decay $\lambda$ values for Boltzmann sampling: (1, 0.1), \textbf{(0.5, 0.05)};
        \item $\alpha$ and $\beta$ prior values for tabular model TS: \textbf{(1, 1)}, (10, 10).            
    \end{enumerate}
    
    All settings were evaluated for 10 iterations (each iteration being 1 human selection, automated emitter steps, and associated MAP-Elites updates, as described in \Cref{alg:emitters_overview}).

    \subsection{Experimental Setup}
        Beyond evaluating different learning-based emitter setups based on our PLE framework, we used the following emitter settings as baselines:
        \begin{enumerate}
            \item the \textit{random emitter}, an emitter with a selection history window of size 0, no model, and uniform sampling; and
            \item the \textit{greedy emitter}, an emitter with a selection history window of size 1, a tabular model, and greedy sampling.
        \end{enumerate}
        The use of the random emitter as a baseline allows us to test whether uniformly selecting from bins to evolve (which is the default setting for MAP-Elites) can produce content that is of interest to the user. The greedy emitter is the most naive implementation of an emitter that mimics user behaviour, and allows us to test whether more complex PLE combinations yield more appealing results.
    

    \subsection{Evaluation}
        We evaluated different system settings according to the following properties:
        \begin{enumerate}
            \item the elapsed time per emitter step;
            \item a qualitative \enquote{alignment level}; and
            \item a qualitative serendipity level,
        \end{enumerate}
        where the latter two were based on the authors' judgement.
        
        For the elapsed time, we measure the entire selection process (feature extraction, predicting the logits $L$, and sampling a bin), and the emitter update. The alignment level is a qualitative metric that we use to determine how closely the emitter's results match the user's expectations\textemdash a measure of how well it has learned the user's preferences. However, we also want the system to generate items that are outside of the user's immediate expectations, but nonetheless may be of interest\textemdash which we measure qualitatively as serendipity. 
        
        The ideal emitter configuration should produce new solutions in a reasonably short time, with a balance between alignment and serendipity.
        

    \subsection{Results}
        The two baselines gave predictable results. The random emitter was very fast but lacked alignment. Its serendipity was highly variable\textemdash depending on chance, sometimes it yielded serendipitous solutions, but mainly it did not. Conversely, the greedy emitter (which also has the benefit of being very fast), had almost perfect alignment, but at the cost of very little serendipity.
        
        For the PLEs, we first checked our models for alignment using a greedy sampling strategy to remove any stochastic effects from sampling. After hyperparameter tuning, we tested them with the more advanced sampling strategies.
        
        We first tested the simple tabular model with the $L$ matrix built as normalised bin counts (\Cref{eq:logits_baseline}), and the more advanced ($\delta$,$\lambda$)-tabular model with the $L$ matrix built using \Cref{eq:nonparametric-logits} instead. The tabular models were the fastest within our PLE framework, and had reasonable alignment. The ($\delta$,$\lambda$)-tabular model had better alignment, but was sensitive to the choice of hyperparameters. The choice for the increment $\delta$ and decay $\lambda$ values altered the results significantly. In particular, a high decay had a similar effect to a smaller window size, as they both limit the model's memory of past user selection, whereas a (proportionally) higher increment resulted in a model that was slower to adapt to changes in user preferences. When tuning window size $k$, we found that keeping all data ($k = \infty$) worked best (together with non-negative decay).
        
        As having a larger window size was beneficial when testing the tabular models, we fixed window size $k = \infty$ for testing the other models. Across the different machine learning models, we found that using the full solution context $S$ as the input features produced better results than just $BC$s.        
        
        We were able to achieve good alignment with the parametric models. The linear models had the benefit of being relatively quick, and we did not find a noticeable difference between linear regression and ridge regression. However, we were able to achieve better alignment with a shallow neural network, at the cost of marginally higher time to update the model. We therefore consider the neural network to be the best, with a favourable trade-off between time and alignment.
        
        The nonparametric models performed worse than the parametric models, both in terms of time and alignment. kNN regression had the best alignment amongst these but was also the slowest, whereas the difference in both time and alignment between linear and nonlinear kernels in KRR was not particularly noticeable. 
    
        Overall, the choice of sampling strategy had a large impact on the qualitative metrics. In almost all settings, Boltzmann sampling with a lower initial $\tau$ and small decay $\lambda$ had a good balance between alignment and serendipity compared to $\epsilon$-greedy. The simple tabular model also allowed us to test TS. Compared to the ($\delta$,$\lambda$)-tabular model with BS, the former performed better in terms of serendipity, but it required more iterations before reaching a good alignment level. However, after a few more iterations, it was also the best model at adapting to shifts in user preferences.

\section{User study}\label{sec:userstudy}

    \noindent After our internal study, we set up a user study with players of Space Engineers. Our applications, including Windows executables, were made open source on GitHub, and the AI Spaceship Generator was advertised by Keen Software House, the developers of Space Engineers, as well as GoodAI, a sister company of Keen Software House, on October 27, 2022.
        
    \subsection{Procedure}
    
\begin{figure}[!b]
        \centering
        \includegraphics[width=.48\textwidth]{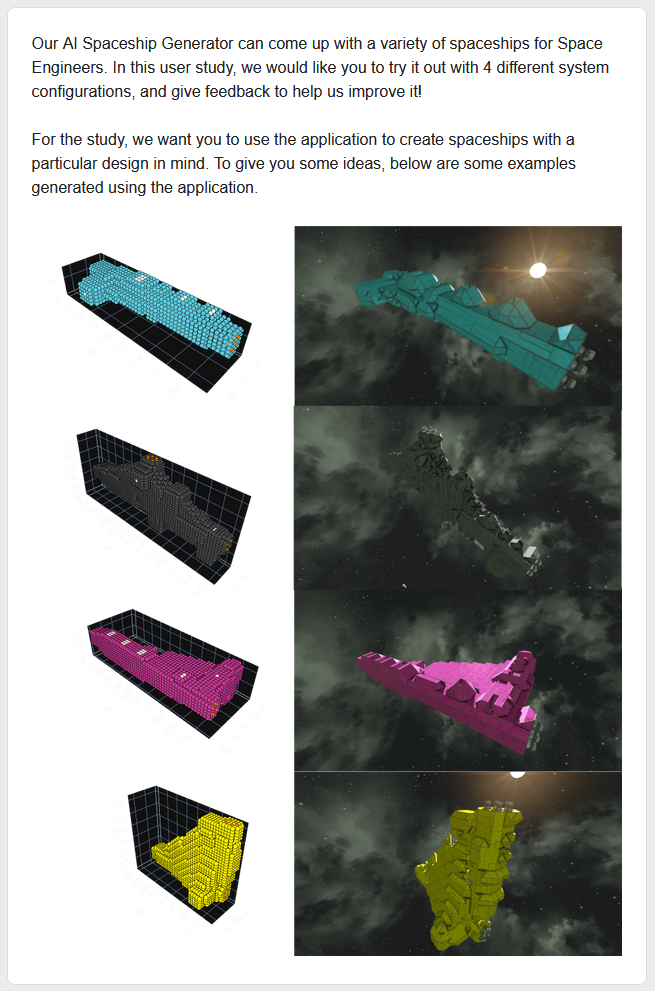}
        \caption{Screenshot of the questionnaire introduction.}
        \label{fig:questionnaire-introduction}
    \end{figure}

\begin{figure}[!t]
        \centering
        \includegraphics[width=.48\textwidth]{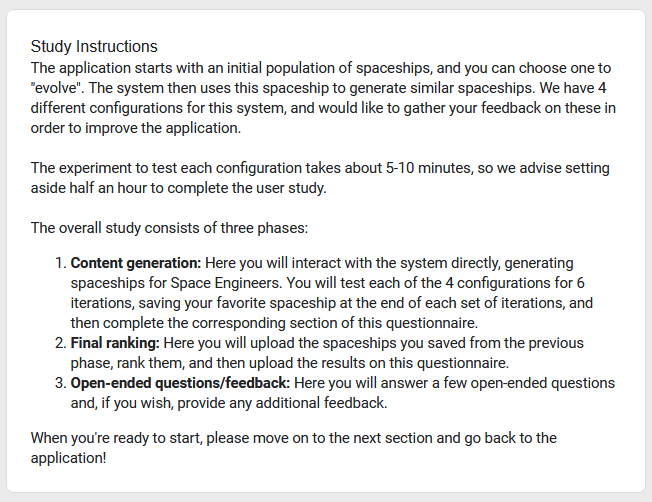}
        \caption{Screenshot of the questionnaire instructions.}
        \label{fig:questionnaire-instructions}
    \end{figure}

    Players were directed towards the releases page, which contained the executables as well as basic instructions. Players were able to download the AI Spaceship Generator and use the normal user mode, or if they agreed to take part in the user study (which required accepting a privacy policy), would enter the application in user-study mode. Launching user-study mode would also open a Google Forms questionnaire with more information about the study (\Cref{fig:questionnaire-introduction}) and instructions (\Cref{fig:questionnaire-instructions}). Users were shown spaceships generated with our system and asked to try and generate their own.
    
    \begin{figure}[!t]
        \centering
        \includegraphics[width=.48\textwidth]{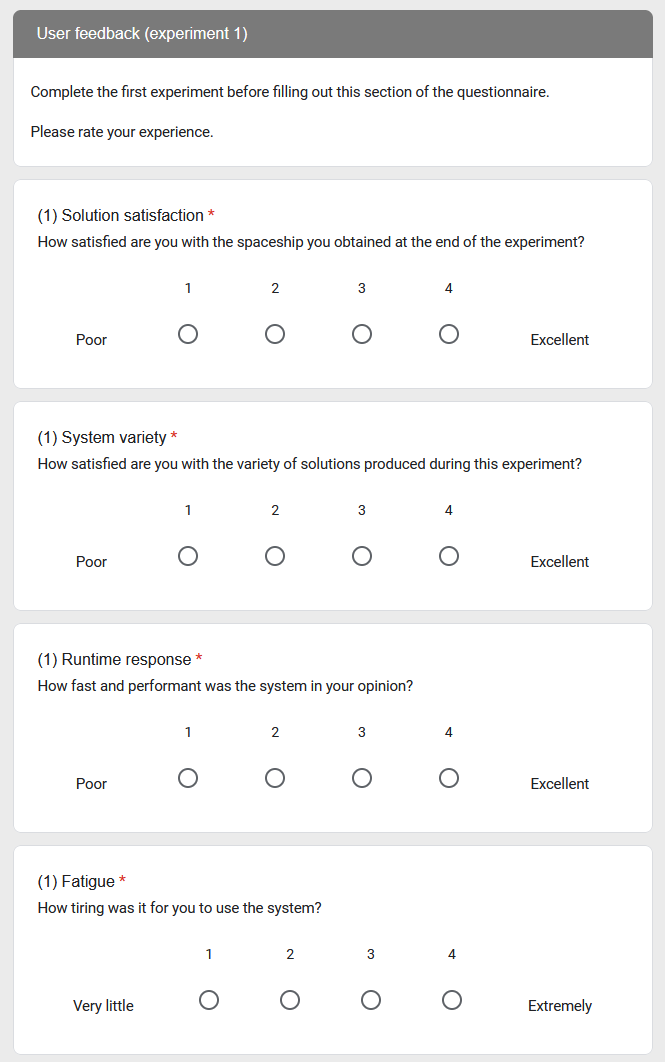}
        \caption{Screenshot of the questionnaire for one configuration. Users assign a vote to each entry after completing the experiment for the current configuration.}
        \label{fig:questionnaire-preview}
    \end{figure}

    After entering a randomised user ID (UID) generated by the application into the form, 
    users took part in 4 experiments, each with a different emitter (ordered randomly and invisibly to the user). 
    The emitters tested were the \enquote{null emitter} (no emitter, only human selections), the random emitter, the greedy emitter, and our best PLE (the neural network with full history and Boltzmann sampling). For each experiment, after 6 iterations of IC MAP-Elites, users were asked to pick their favourite spaceship from the final generation (which was then automatically downloaded), and then fill out the relevant section of our questionnaire (\Cref{fig:questionnaire-preview}).
     
    At the end of all the experiments, the user was then guided to download and launch the separate Spaceships Ranker application. After uploading the 4 spaceships chosen during the experiments, the user was asked to rank them subjectively, and upload the ranking on the questionnaire. Finally, users were invited to give free-form feedback.

    \subsection{Results}

    At the end of the user study (October 27 through November 28, 2022), we received feedback from 95 users, of which only 52 were valid.\footnote{The other 43 results were discarded due to either incomplete feedback, or wrong or duplicate files submitted} The data was anonymised before storage and analysis.
    
\begin{figure}[!h]
        \centering
        \includegraphics[width=0.2522\textwidth,valign=t]{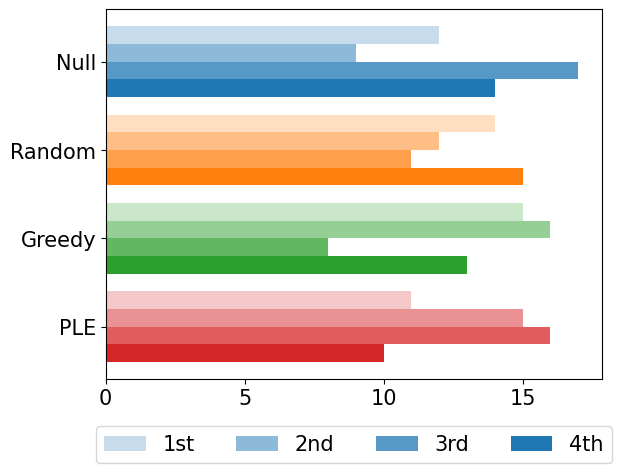}
        \includegraphics[width=0.23\textwidth,valign=t]{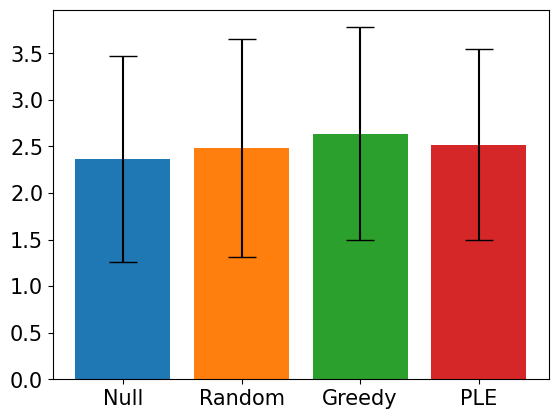}
        \caption{Emitter rankings. (Left): Number of times each emitter was ranked in each position. (Right): Average emitter scores ($\pm 1$ standard deviation).}
        \label{fig:emitters-stats}
    \end{figure}

    We report the rankings and average scores ($1^\text{st}$ place scoring 4 points, $2^\text{nd}$ scoring 3 points, etc.) in \Cref{fig:emitters-stats}. The random and greedy emitters were ranked first place most often (14 and 15 times, respectively), but the random emitter was also ranked last place the most alongside the human emitter (15 and 14 times, respectively). The PLE was ranked fourth place the least (10 votes), and mostly ranked second or third place (15 and 16 times).
    
    In order to proceed with a statistical analysis of the ranking data, we applied the Shapiro-Wilk test \cite{shapiro_analysis_1965} to choose an appropriate statistical method. With a test statistic $W \approx 0.84$ for all emitters ($p < 0.0001$), the ranking data was found to be not normally distributed. Therefore, we used the Friedman test (non-parametric ANOVA) \cite{friedman_1940} to check whether the average scores were significantly different. With $\chi^2 = 1.15$ ($p = 0.764$), we conclude that there was no major difference on the \emph{final} spaceships evaluated. We believe there are at least two reasons why there was no clear favourite configuration:
    \begin{enumerate}
        \item The solutions generated in this domain were not too different from each other. This is reflected in the user feedback we received: \enquote{\textit{i hope to see more variety in the ships\ldots}}, \enquote{\textit{\ldots the current implementation does not result in very interesting ships, more like a couple of variations of a fairly similar shape}}, \enquote{\textit{\ldots also all the ships are just spread out from a central design and are very limited in actual variations}}, \enquote{\textit{Looking forward to seeing the designs get more diverse\ldots}}; and
        \item The amount of iterations was too high for this domain, resulting in the system finding similar solutions across configurations, as noted in user feedback: \enquote{\textit{I basically kept getting the same shapes of ships as I was picking what looked good\ldots}}, \enquote{\textit{\ldots the last 3-4 iterations (of every experiment) are mostly useless, since a good 80-90\% of the best ships are produced in the first two iterations\ldots}}.
    \end{enumerate}
    For the above reasons, we believe it may be possible to see greater benefits of the PLE framework in a less constrained search space.

    \begin{figure}[!h]
        \centering
        \includegraphics[width=0.23\textwidth]{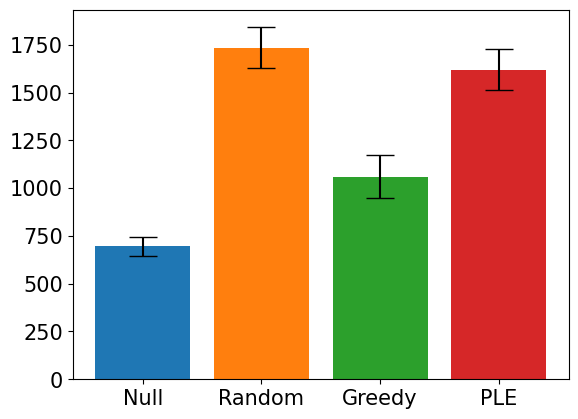}
        \includegraphics[width=0.23\textwidth]{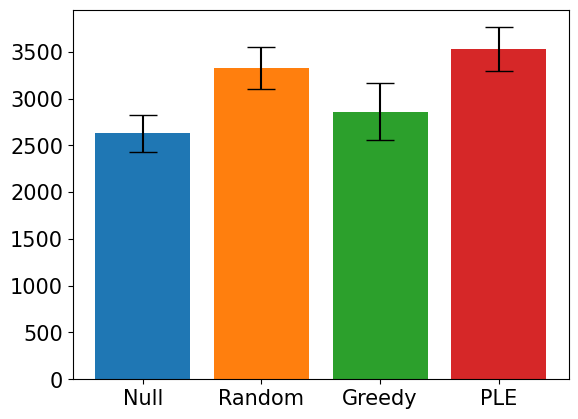}
        \\
        \includegraphics[width=0.23\textwidth]{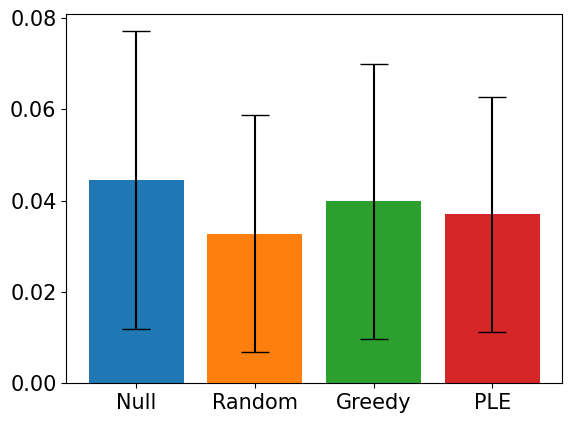}
        \includegraphics[width=0.23\textwidth]{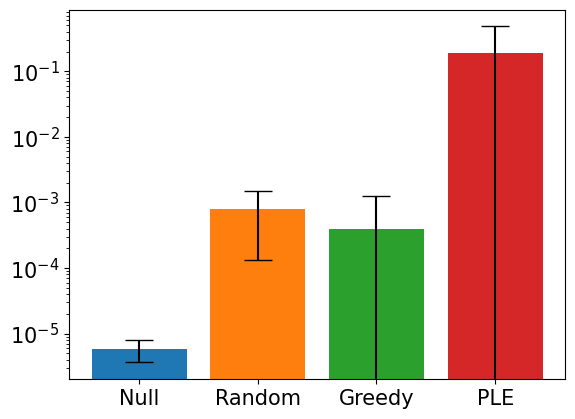}
        \caption{Application statistics (mean $\pm$ 1 standard deviation, calculated over users and across iterations of IC MAP-Elites). (Top left): Number of solutions generated. (Top right): Complexity (L-system string length) of the generated solutions. (Bottom left): Number of spaceships inspected, normalised by population size. (Bottom right): Time taken (in seconds, displayed in log scale) to complete an emitter step, averaged over all generations.}
        \label{fig:user-metrics}
    \end{figure}

    We also report different metrics collected during the user study in \Cref{fig:user-metrics}. A clear trend is that the random emitter and PLE are able to better explore the solution space, resulting in more diverse populations. The \enquote{complexity}\footnote{Computed as the length of the L-system string that defines the spaceship; we also put a constraint on the maximum length to avoid spaceships that are invalid due to game constraints.} of their solutions are also higher. The number of spaceships inspected by users was roughly proportional to the population size, and hence there was no significant difference between the emitters when accounting for this. Finally, the PLE took significantly longer than the other emitters, but only on the order of one tenth of a second.
    
    Additionally, as is common in simple user studies \cite{gonsalves_user_2014}, users were asked to rate various properties of the configuration on a 4-item Likert (ordinal) scale: the satisfaction with the final solution obtained, the population variety, the real-time response, and the fatigue level at the end of each interaction with the system. The responses to this questionnaire are reported in \Cref{tab:questionnaire-res}. Similarly to the emitter rankings, we converted the results into a point-based system to derive a final score for each item, from 1 to 4 for \enquote{Poor} to \enquote{Excellent}. The Shapiro-Wilk test statistic $W$ ranged from $0.4\textrm{--}0.9$ for most features ($p < 0.001$). We therefore performed the Friedman test on the 4 metrics from our questionnaire. The only metric with statistical significance was the \enquote{runtime response} with $\chi^2 = 23.0$ ($p < 0.001$). We then performed post-hoc analysis using the Durbin-Conover test \cite{conover_practical_1999} for pairwise comparisons on this metric: both the Null and Greedy emitters scored significantly higher than the PLE ($T = 4.75$ and $3.48$ respectively, $p < 0.001$ after applying the Bonferroni correction).
    
    
    \begin{table}[h]
        \centering
        \resizebox{.48\textwidth}{!}{%
        \begin{tabular}{|c|c|cccc|c|}
        \hline
        \multicolumn{1}{|c|}{\multirow{2}{*}{\textbf{Emitter}}} & \multirow{2}{*}{Metric} & \multicolumn{4}{c|}{Count} & \multirow{1}{*}{Score} \\
        \multicolumn{1}{|c|}{} &  & Poor & Fair & Good & Excellent & (Average)  \\ \hline
        \multirow{4}{*}{Null} & Solution Satisfaction & 5 & 16 & 19 & 12 & 2.73 \\
         & System Variety & 20 & 13 & 10 & 9 & 2.15 \\
         & Runtime Response & 5 & 19 & 22 & 6 & 2.56 \\
         & Fatigue & 8 & 26 & 13 & 5 & 2.29 \\ \hline
        \multirow{4}{*}{Random} & Solution Satisfaction & 4 & 14 & 26 & 8 & 2.73 \\
         & System Variety & 12 & 16 & 14 & 10 & 2.42 \\
         & Runtime Response & 13 & 19 & 18 & 2 & 2.17 \\
         & Fatigue & 10 & 22 & 14 & 6 & 2.31 \\ \hline
        \multirow{4}{*}{Greedy} & Solution Satisfaction & 7 & 10 & 27 & 8 & 2.69 \\
         & System Variety & 14 & 19 & 10 & 9 & 2.27 \\
         & Runtime Response & 9 & 15 & 24 & 4 & 2.44 \\
         & Fatigue & 13 & 22 & 13 & 4 & 2.15 \\ \hline
        \multirow{4}{*}{PLE} & Solution Satisfaction & 4 & 10 & 27 & 11 & 2.87 \\
         & System Variety & 11 & 17 & 16 & 8 & 2.40 \\
         & Runtime Response & 16 & 20 & 12 & 4 & 2.08 \\
         & Fatigue & 15 & 19 & 13 & 5 & 2.15 \\ \hline
        \end{tabular}%
        }
        \caption{Questionnaire results: user feedback in response to the different system configurations.}
        \label{tab:questionnaire-res}
    \end{table}

     Finally, we are pleased to report that the vast majority of feedback the application received was positive, with most users stating that the generator was a great starting point for generating spaceships, \enquote{\textit{seeing great potential}} in it and \enquote{\textit{I enjoyed} [it] \textit{and got some ideas for new ships}}. The application was also generally perceived as \enquote{\textit{easy to use}}. Many users also had feature requests, which we have since then implemented for the normal user mode.
     

\section{Conclusion}\label{sec:conclusions}
    
    
    \noindent In this paper we introduced the PLE framework, incorporating preference learning into the mixed-initiative co-creation PCG setting with QD algorithms. While QD algorithms offer an attractive approach for co-creation by illuminating a diverse set of solutions, by default the user might need many interactions with the system in order to create a desirable solution. By modelling user preferences and performing automated updates using the emitter framework, we are able to greatly improve the experience for the user.
    
    We validated our PLE framework on a PCG task for the Space Engineers video game using a qualitative internal study. In order to achieve this, we further extended prior work in this domain \cite{gallotta_evolving_2022,gallotta_surrogate_2022} by improving the spaceship generation algorithm and making the system interactive with an application interface so that it can be used in the mixed-initiative setting.

    We also conducted a user study with Space Engineers players, and collected useful feedback and metrics. Whilst the use of emitters definitely improved the size of the solution space that could be explored by users, there was no clear winner when users were asked to rank their final solutions for each emitter configuration.
    
    A strength of PLEs is the general framework around them that we have developed. This provides many directions for future research, adapting different models or different sampling strategies. Preliminary results using model ensembles were promising, making this a fruitful avenue to pursue.
    
    
    
    Considering the goal of improving the user experience in co-creation, another future research direction would be to extend PLEs to incorporate meta-information in the selection process (for example, additional performance metrics, such as the number of solutions generated) to further improve the alignment and serendipity we could obtain with the system.

\section*{Acknowledgements}
    \noindent This project was partly funded by a GoodAI research grant. The authors would like to thank Martin Poliak, Senior Research Scientist at GoodAI, and Erin Truitt, Community Manager at Keen Software House, for their invaluable feedback on the application. The authors would also like to thank Francesco Crottini for his feedback on the user study analysis process, Space Engineers player \emph{Bardaky} for the continuous feedback on the early stages of the application, and the Space Engineers community for using our application and participating in the user study.

\bibliographystyle{acm}
\bibliography{references}







\end{document}